\documentclass[lettersize,journal]{IEEEtran}
\usepackage{amsmath,amsfonts}
\usepackage{algorithm}
\usepackage{array}
\usepackage{textcomp}
\usepackage{stfloats}
\usepackage{url}
\usepackage{verbatim}
\usepackage{graphicx}
\usepackage{cite}
\usepackage{amssymb}
\usepackage{xcolor}
\usepackage{multirow}
\usepackage{colortbl}
\usepackage{makecell}
\usepackage{algorithmicx}
\usepackage[caption=false,font=footnotesize,labelfont=sf,textfont=sf]{subfig}

\newtheorem{theorem}{Theorem}

\newcommand{\wqza}[1]{{\color{black}#1}}
\newcommand{\myref}[1]{{\color{black}#1}}

\begin{document}

\title{Dual-CBA: Improving Online Continual Learning\\via Dual Continual Bias Adaptors\\from a Bi-level Optimization Perspective}

\author{Quanziang Wang, Renzhen Wang, Yichen Wu, Xixi Jia, Minghao Zhou, Deyu Meng~\IEEEmembership{Member,~IEEE,}
\thanks{Quanziang Wang, Renzhen Wang, Minghao Zhou, and Deyu Meng are with the School of Mathematics and Statistics, Xi'an Jiaotong University, Xi'an, Shaanxi, P.R.China (email: quanziangwang@gmail.com; ~rzwang@xjtu.edu.cn;~~woshizhouminghao@stu.xjtu.edu.cn;~~dymeng@mail.\\xjtu.edu.cn).}
\thanks{Yichen Wu is with the Department of Computer Science, City University of Hong Kong, Tat Chee Avenue, Kowloon Tong, Hong Kong (email: wuyichen.am97@gmail.com).}
\thanks{Xixi Jia is with the School of Mathematics and Statistics, Xidian University, Xi'an, Shaanxi, P.R.China (email: hsijiaxidian@gmail.com).}}

\markboth{Journal of \LaTeX\ Class Files,~Vol.~14, No.~8, August~2021}%
{Shell \MakeLowercase{\textit{et al.}}: A Sample Article Using IEEEtran.cls for IEEE Journals}


\maketitle

\begin{abstract}
In online continual learning (CL), models trained on changing distributions easily forget previously learned knowledge and bias toward newly received tasks. To address this issue, we present Continual Bias Adaptor (CBA), a bi-level framework that augments the classification network to adapt to catastrophic distribution shifts during training, enabling the network to achieve a stable consolidation of all seen tasks. However, the CBA module adjusts distribution shifts in a class-specific manner, exacerbating the stability gap issue and, to some extent, fails to meet the need for continual testing in online CL. To mitigate this challenge, we further propose a novel class-agnostic CBA module that separately aggregates the posterior probabilities of classes from new and old tasks, and applies a stable adjustment to the resulting posterior probabilities. We combine the two kinds of CBA modules into a unified Dual-CBA module, which thus is capable of adapting to catastrophic distribution shifts and simultaneously meets the real-time testing requirements of online CL. Besides, we propose Incremental Batch Normalization (IBN), a tailored BN module to re-estimate its population statistics for alleviating the feature bias arising from the inner loop optimization problem of our bi-level framework. To validate the effectiveness of the proposed method, we theoretically provide some insights into how it mitigates catastrophic distribution shifts, and empirically demonstrate its superiority through extensive experiments based on four rehearsal-based baselines and three public continual learning benchmarks.
\end{abstract}


\section{Introduction}
\label{sec:introduction}
\IEEEPARstart{C}{ontinual} learning (CL)~\cite{cl-settings, cl-survey} aims to develop models that can accumulate new knowledge while consolidating previously learned knowledge from streaming data. In the context of CL, the data distribution of streaming tasks is in general non-stationary and changes over time, which violates the independent and identically distributed (i.i.d) assumption commonly adopted in traditional machine learning. Therefore, continual learning suffers from catastrophic forgetting problem~\cite{forgetting}, where the model severely forgets the previously learned knowledge after being trained on a new task.

Traditional offline CL stores all training batches of the current task and the model is trained on these samples for multiple epochs to achieve relatively superior performance. However, the availability of previously learned batches might be restricted due to privacy concerns~\cite{cl-survey2} or memory limitations. In this paper, we mainly focus on online CL~\cite{online-survey}, a more challenging and realistic setting. In online CL, samples from each task can be trained in only a single-pass (\textit{i.e.}, one epoch), and previous batches are not accessible in the future. 

Unlike traditional CL settings, the training data distribution in online CL continuously changes throughout the entire training process. Consequently, online CL often leads to more severe distribution shifts, further exacerbating catastrophic forgetting. To alleviate this problem, rehearsal-based methods~\cite{icarl, DER, RAR, cl-survey} employed a small memory buffer to store the examples of previous tasks, aiming at approximating the entire data distribution of all seen tasks. Even though these rehearsal-based methods have achieved sound performance in online CL, most of them often suffer from task-recency bias~\cite{SCR}, \textit{i.e.}, the classifiers tend to classify samples into the classes that are currently being trained. Consequently, some previous works aim to improve the original linear classifier~\cite{SSIL, LUCIR, bic} or replace it directly with the nearest classifier~\cite{SCR, icarl} to mitigate the negative effects of class imbalance between currently received classes and replayed classes. Despite the promising performance, almost all of these methods implicitly view task-recency bias as a label distribution shift and tackle it from the perspective of class imbalance, which makes these methods sub-optimal in practice \cite{online-er-ace}.

\begin{figure}
\centering
\includegraphics[width=0.85\columnwidth]{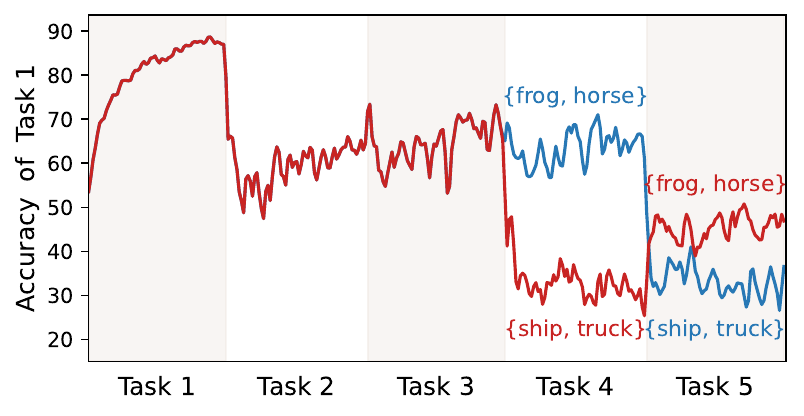}
\caption{Tracking the accuracy of the 1st task with different incoming classes of the 4th and 5th tasks as plotted by red and blue lines, respectively. The label distribution $\mathbb P(Y)$ remains unchanged between the two lines, while the final accuracy of the 1st task varies dramatically. This indicates the effectiveness of the feature distribution shifts (\textit{i.e.}, changes of $\mathbb P(X|Y)$) in CL.}
\label{fig:discrepancy}
\vspace{-2mm}
\end{figure}

From the Bayes viewpoint, the target of online CL is to accomplish a stable consolidation of knowledge for all tasks by fitting the posterior probability $\mathbb P(Y|X)$, where $X$ and $Y$ represent stochastic variables of the input data and the corresponding label, respectively. According to Bayes's rule $\mathbb P(Y|X) \propto \mathbb P(X|Y)\mathbb P(Y)$, any shift from the prior probability $\mathbb P(Y)$ or the likelihood $\mathbb P(X|Y)$ will lead to distribution change in $\mathbb P(Y|X)$. For example, as previous works point out~\cite{SSIL, LUCIR, bic, online-er-ace}, incoming new tasks can change the label distribution $\mathbb P(Y)$, leading to severe forgetting. On the other hand, $\mathbb P(X|Y)$ can also suffer from catastrophic shifts (dubbed feature distribution shift for simplicity) due to the time-varying data streams, especially in online continual learning (CL). To illustrate the feature distribution shift, we conducted a toy experiment using Experience Replay (ER)~\cite{ER1} on CIFAR-10~\cite{cifar}. As shown in \myref{Fig.~\ref{fig:discrepancy}}, we exchange the incoming classes of the 4th and 5th tasks while maintaining the label distribution $\mathbb P(Y)$ unchanged, as plotted in red and blue lines, respectively. By continuously tracking the accuracy of the 1st task, we observe a significant difference in its final accuracy. This validates the existence of feature distribution shifts (\textit{i.e.}, changes in $\mathbb P(X|Y)$) and highlights a challenging problem for online continual learning (CL): \textit{how to achieve stable consolidation of past knowledge amidst these distribution shifts.}

To tackle these challenges, we introduce the Continual Bias Adaptor (CBA) module to directly adapt the posterior distribution shift online in our conference version~\cite{CBA}. This module aids the original classification network in aligning with the evolving posterior distribution, thereby facilitating stable knowledge consolidation across all seen tasks. To jointly optimize the classification network and the CBA module, we propose a bi-level optimization framework, in which the inner loop problem optimizes the classification network by rehearsal-based methods with the help of CBA, and the outer loop optimizes CBA to assimilate the training bias in continual learning. Specifically, the CBA module in~\cite{CBA} is designed as a lightweight neural network such as a multi-layer perceptron (MLP), which maps the posterior distribution predicted by the classification network to the corresponding adapted posterior probability in a class-specific manner. For clarity, we refer to the CBA module proposed in the conference version as the class-specific CBA in the subsequent sections.

\begin{figure}
\centering
\includegraphics[width=0.9\columnwidth]{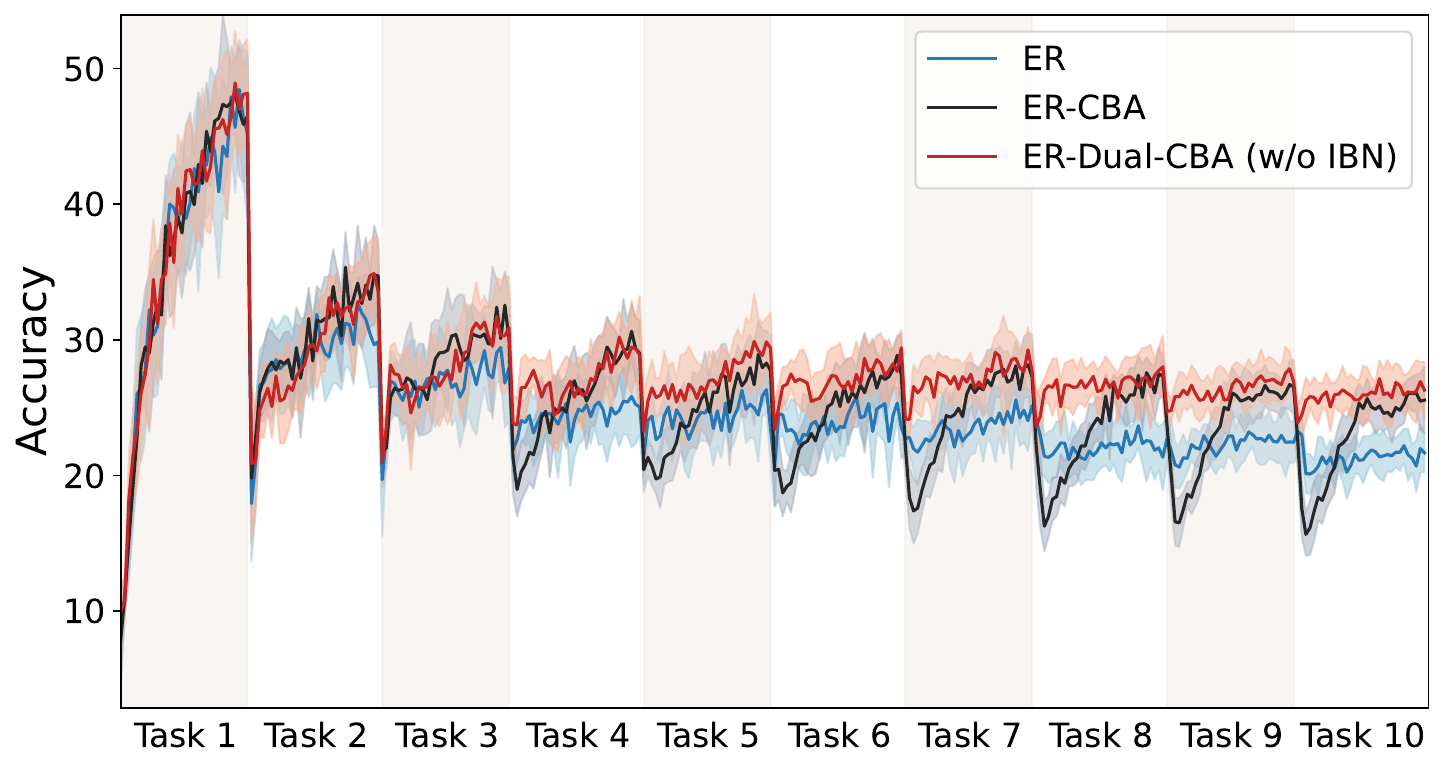}
\caption{The average accuracy during the whole continual training process on Split CIFAR-100 with memory buffer size M=2k. The stability gap problem means the performance of old tasks drops upon starting to learn a new task and then recovers quickly. Comparing the baseline ER, the method ER-CBA presented in our conference version~\cite{CBA} aggravates the stability gap while our ER-Dual-CBA alleviates this problem effectively.}
\label{fig:acc-auc-cifar100}
\vspace{-2mm}
\end{figure}

Albeit fast adaptation to drastic posterior distribution change, the class-specific CBA module suffers from a serious stability gap problem~\cite{stability-gap}, where the performance of the previously learned tasks significantly drops upon starting to learn a new task, followed by a fast recovery phase. To illustrate this phenomenon, we compare the accuracy curve of the baseline model ER with that of the class-specific CBA (ER-CBA) on CIFAR-100 in \myref{Fig.~\ref{fig:acc-auc-cifar100}}. It can be observed that ER-CBA achieves higher performance at the end of each task training, but it suffers a more substantial degradation than ER upon starting to train a new task. This indicates that the class-specific CBA cannot immediately adapt to the catastrophic posterior distribution change caused by incoming new tasks, thus compromising the need for continuous performance evaluation at any time in online CL. This paper reveals that this serious stability gap arises from the class-specific CBA's input, specifically the posterior distribution predicted by the classification network, which changes dramatically upon starting to learn a new task or at the task transition timestamp. Thus, the class-specific CBA module may produce incorrect adjustments for the classification network, resulting in a substantial stability gap.

To mitigate the stability gap problem, we propose a novel class-agnostic CBA designed to capture the stable relationship between the new task and all previously learned tasks in the continual learning \wqza{process}. The proposed class-agnostic CBA aggregates the overall posterior probabilities of the new task and those of all the old tasks separately, subsequently yielding stable corrections for the resulting posterior distribution. As a robust and transferable bias adaptor, the class-agnostic CBA module can effectively adapt to sudden changes in posterior distributions and thus can mitigate the stability gap problem. To further leverage the advantages of both the class-agnostic CBA and the class-specific CBA, we integrate these two modules into a unified Dual-CBA.
This integration allows the classification network to effectively fit the implicit posterior distribution while maintaining stable knowledge consolidation from all seen tasks throughout the entire continual learning process, as shown in \myref{Fig.~\ref{fig:acc-auc-cifar100}}.

Furthermore, we found that the bi-level framework proposed in \cite{CBA} encounters a feature bias challenge. This challenge arises from the commonly used batch normalization (BN)~\cite{BN}, which induces feature deviations during the testing \wqza{stage}. Specifically, the population statistics of BN are only updated in the inner loop together with the classification network parameter according to the samples from the new task and those saved in the memory buffer. This causes the exponential moving average (EMA) of BN used for testing to be gradually dominated by the new task because most of the training data in the inner loop comes from the new tasks, leading to biased population statistics. Unfortunately, this bias is not fed back into the posterior distribution, making it challenging for the proposed Dual-CBA to adapt accordingly.

To address this problem, we propose Incremental BN (IBN), a simple yet effective method to mitigate the aforementioned feature bias. IBN stops the updating of EMA population statistics of BN in the inner loop and instead replaces them with statistics calculated from the outer loop. This strategy yields a balanced population estimation for the classification model and effectively assimilates the feature bias for the testing stage. For a simple implementation, we can only update the population statistics before the testing stage with the memory buffer data, which shows significant performance improvement for the proposed bi-level learning algorithm.

In a nutshell, building on the class-specific CBA published in our conference version~\cite{CBA}, this paper extends it from several perspectives. The main contributions are summarized as follows,
\begin{itemize}
    \item We shed light on certain limitations of our previously designed class-specific CBA module. On this basis, we propose the class-agnostic CBA and combine it with the class-specific CBA to form a novel Dual-CBA, which can effectively assimilate the training bias and alleviate the stability gap issue in continual learning.
    \item Based on the proposed bi-level learning framework, we propose Incremental BN to estimate more accurate BN statistics with a simple implementation, assimilating the feature bias during testing.
    \item Theoretically, we explain how the proposed method adapts to distribution shifts and alleviates forgetting in CL from the perspective of gradient alignment. Additionally, we provide insights into the optimization process of the bi-level learning framework from a linear formulation.
    \item We conduct extensive experiments to demonstrate that our method consistently improves upon various rehearsal baselines across multiple continual learning (CL) settings. Specifically, we extend Dual-CBA to a semi-supervised continual learning setting and further validate its generalization ability.
    \item We highlight the strong transferability of the proposed class-agnostic CBA module to unseen tasks or datasets when pre-trained on a limited number of tasks. Concretely, we demonstrate that a pre-trained module can be directly applied to various intra-dataset and inter-dataset scenarios.
\end{itemize}
The paper is organized as follows. \myref{Sec.~\ref{sec:related-works}} discusses related works. \myref{Sec.~\ref{sec:method}} introduces the proposed Dual-CBA method in detail. \myref{Sec.~\ref{sec:experiments}} present extensive experiments to evaluate our method, and the conclusions are summarized in \myref{Sec.~\ref{sec:conclusion}}.

\section{Related Works}
\label{sec:related-works}

\noindent\textbf{Continual learning settings.}
Based on different task construction manners, continual learning (CL) mainly falls into three categories~\cite{cl-settings, cl-survey, cl-survey3}: Task-incremental learning (Task-IL), Domain-incremental learning (Domain-IL), and Class-incremental learning (Class-IL). Specifically, Task IL necessitates prior knowledge of the task index in both the training and testing stages. Domain-IL mainly focuses on concept drift, where the domain of each task changes while the label space remains unchanged~\cite{domain-il3, domain-il2, domain-il1}. This paper concentrates on the more challenging Class-IL, where the task index is unavailable during testing~\cite{DER, class-il1, icarl, class-il2}. Additionally, from the training perspective, CL can be divided into offline and online CL. Offline CL involves preserving all samples of the current task and training the model on them across multiple epochs~\cite{CLSER, DER, e2e, offline1, icarl}. As for online CL, samples of the current task arrive sequentially, which cannot be stored entirely, and each sample is typically seen once, except when stored in the memory buffer~\cite{ACC_AUC, online-survey}. In this paper, we mainly focus on online CL, which represents a more demanding and realistic setting compared to offline CL. Furthermore, since it is expensive to obtain a large amount of labeled data, many works focus on certain weakly supervised scenarios, such as few-shot~\cite{cl-few-shot1, cl-few-shot2, cl-few-shot3}, semi-supervised~\cite{cl-semi-sup1, cl-semi-sup2, cl-semi-sup3}, and imbalance~\cite{online-CL-imbalance, imbalance-CL}, etc. To demonstrate the flexibility of our approach, we also extend our method to the semi-supervised continual learning setting.

\medskip
\noindent\textbf{Rehearsal-based methods in online CL.}
In online CL, the main objective is to make the model quickly acquire new knowledge from a new task while retaining previously learned knowledge from old tasks~\cite{online-er-ace, online-imbalance, topology1, topology2, L2Imagine, ACC_AUC, RER}. A commonly used baseline, Experience replay (ER), trains the incoming new samples along with old samples from the memory buffer together. The variants of ER attempt to employ different techniques in the replay strategy, such as knowledge distillation and random augmentation. For example, DER++~\cite{DER} utilized a stronger distillation method to further replay logits of the memory buffer data and  Mnemonics~\cite{mnemonics} applied bi-level optimization to distillate global information of all seen examples into a few learnable ones. RAR~\cite{RAR} adopted random augmentation to alleviate overfitting of the memory buffer, while CLSER~\cite{CLSER} constructed plastic and stable models to consolidate recent and structural knowledge distillation. Different from these methods, some studies emphasize maximizing the utility and benefits of the memory buffer samples~\cite{gradient-selection-blurry, MIR, OCS, data-compression-wly, data-compression-sqr, mnemonics}. For example, instead of randomly sampling, GSS~\cite{gradient-selection-blurry} selected the samples stored in the memory buffer according to the cosine similarity of gradients, MIR~\cite{MIR} chose maximally interfering samples whose prediction will be most negatively impacted by the foreseen parameters update, and OCS~\cite{OCS} picked the most representative data of the current task while minimizing interference with previous tasks. Unlike these methods, our method focuses on alleviating distribution shifts and can plug in most current rehearsal-based approaches.

\medskip
\noindent\textbf{Task-recency bias in CL.}
Task-recency bias \cite{LUCIR, cl-survey} in online CL refers to the tendency of classifiers to mistakenly classify examples belonging to previously learned classes as newly received ones. \wqza{Typically, the linear classifier is susceptible to task-recency bias.} To address this, iCaRL~\cite{icarl} proposed to replace the linear classifier with nearest class mean (NCM) classifiers. Similarly, SCR~\cite{SCR} and Co$^2$L~\cite{Co2L} employed the NCM classifier, where the feature extractor was trained using a contrastive learning paradigm. A wide range of works tackles the task-recency bias as a class imbalance problem~\cite{SSIL, e2e, LUCIR, bic, WA}. For example, LUCIR~\cite{LUCIR} introduced weight normalization to the linear classifier, BiC~\cite{bic} proposed a bias correction layer turned on a held-out validation set, and SS-IL~\cite{SSIL} modified the softmax to mitigate the imbalanced penalization for the outputs of old classes. On the flip side, ER-ACE~\cite{online-er-ace} pointed out that task-recency bias can also arise from feature interference and designed an asymmetric loss to address this problem. Different from these methods, our proposed method relaxes the assumption on label/feature distribution shift by directly modeling the posterior distribution shift.
Besides, some methods addressing class imbalance may potentially be explored for tackling task-recency bias in online CL~\cite{autobalance, logit_adjustment, meta_modulator, L2AC}. However, many of these approaches face challenges in generalization to CL due to unstable data distribution.

\medskip
\noindent\textbf{Normalization Layer in CL.}
The normalization layer is a crucial component of deep neural networks. Batch Normalization (BN)~\cite{BN} has shown a strong performance and become the most commonly used normalization strategy in single-task scenarios with fixed data distribution~\cite{understand_BN1, understand_BN2}. However, BN may hinder the continual learning performance because of evolving data distributions~\cite{CN, TBBN, AdaB2N}. To address this, Continual Normalization (CN)~\cite{CN} reduced the cross-task differences by Group Normalization~\cite{GroupNorm} and subsequently utilized BN to normalize the input features. BNT~\cite{BN-zmh} constructed a balanced batch to update BN statistics. TBBN~\cite{TBBN} addressed the imbalance problem in BN by employing reshape and repeat operations to construct a task-balanced batch during training. \wqza{Task-Specific BN~\cite{task-specific-BN} focused on task-IL and applied specific BN statistics for each task.} Besides, AdaB$^2$N~\cite{AdaB2N} leveraged a regularization term to optimize a modified momentum to balance BN statistics. Different from these works, our proposed IBN is designed to address the feature bias that arises from the bi-level optimization used in this paper.

\section{Method}
\label{sec:method}

\subsection{Preliminaries of Continual Learning}
\label{subsec:preliminaries}
In continual learning, the model is trained on a stream of tasks with evolving data distribution. Considering $N$ sequential tasks $\{\tau_1, \tau_2, \cdots, \tau_N\}$, each task can be represented as $\tau_t=\{(x^t_i,y^t_i)\}_{i=0}^{N_t}$, where $N_t$ is the total number of training samples in task $\tau_t$. For the online CL setting, the classification model $f_\theta$ trains each new sample only once, and the previously seen batches are almost inaccessible. Let $\tau_t$ represent the current learning task, and $\mathcal{M}$ denotes a small memory buffer that stores the samples of previous tasks $\{\tau_{1}, \cdots, \tau_{t-1}\}$. Here we take a widely used rehearsal-based method, Experience Replay (ER)~\cite{ER1, ER2}, as an example. ER jointly trains current task $\tau_t$ with examples sampled from the buffer $\mathcal{M}$. The training objective function $\mathcal{L}$ is
\begin{equation}
    \mathcal{L}^{trn} \left(\mathcal{B}^{trn}; f_\theta \right) = 
    \frac{1}{|\mathcal{B}^{trn}|}
    \sum_{x,y \in \mathcal{B}^{trn}} L(f_\theta(x), y) ,
\end{equation}
where the training batch $\mathcal{B}^{trn}=\mathcal{B}^t \cup \mathcal{B}^{buf}$ consists of a batch of incoming new samples $\mathcal{B}^t \subset \tau_t$ and a batch of samples from the memory buffer $\mathcal{B}^{buf} \subset \mathcal{M}$, and $L$ denotes the cross-entropy loss. Note that the memory buffer $\mathcal{M}$ is updated by reservoir sampling after training each batch $\mathcal{B}_{t}$, which is a relatively balanced set containing samples of all seen tasks.

\begin{figure*}[t]
\begin{center}
   \includegraphics[width=0.95\linewidth]{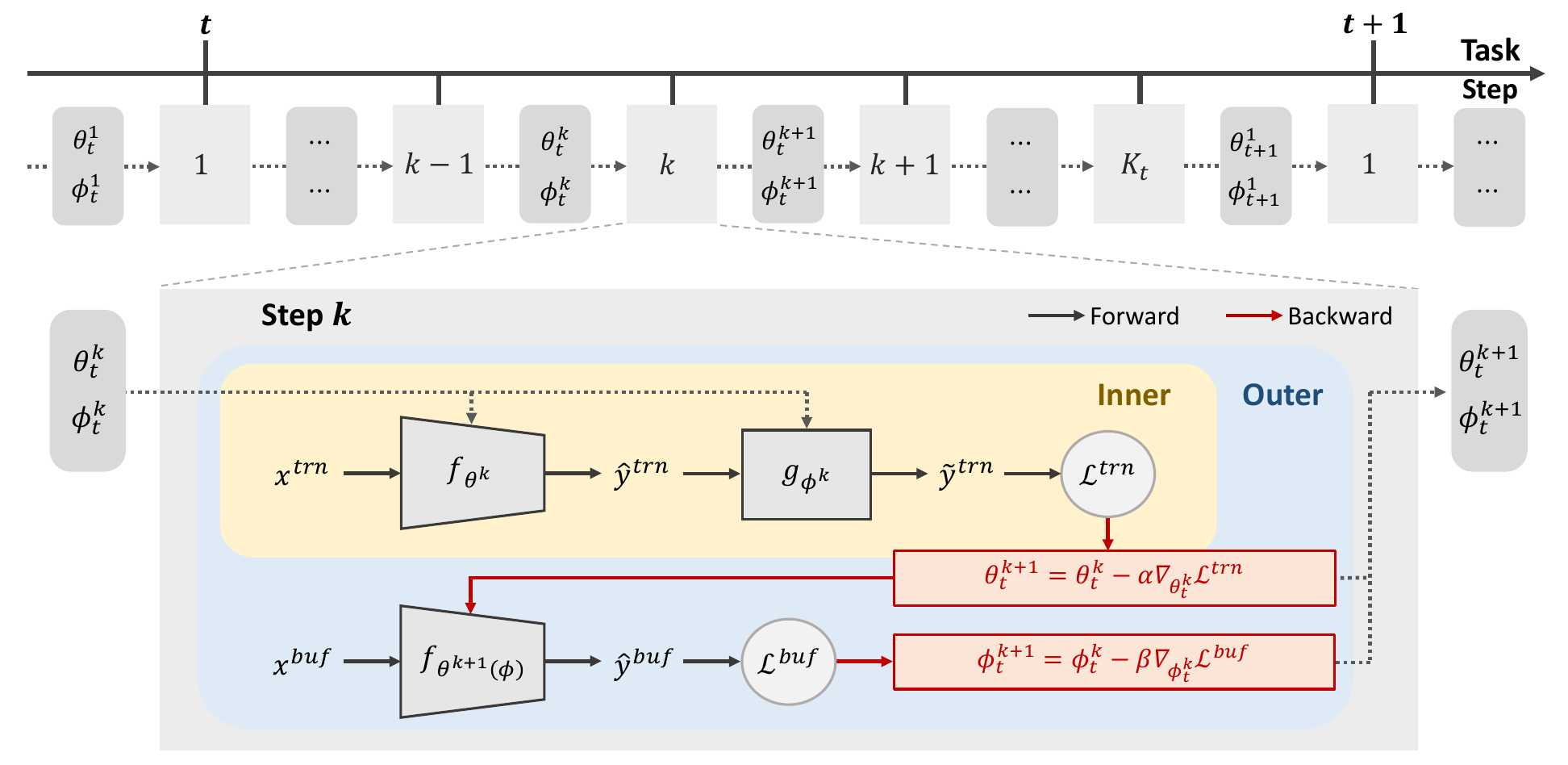}
\end{center}
\vspace{-5mm}
\caption{Method overview. At each iteration step $k$, the classification model parameter $\theta$ and the Dual-CBA parameter $\phi = \{ \omega, \nu \}$ are jointly updated through the bi-level optimization framework, where $\omega$ and $\nu$ represent parameters of the class-specific CBA and class-agnostic CBA, respectively. For the inner loop, the forward process computes the rehearsal training loss and the backward process updates the classification model parameter $\theta(\phi)$ by \myref{Eq.~(\ref{eq:inner-opt})}. For the outer loop, the forward process computes the outer objective loss function and the backward process updates the Dual-CBA parameter $\phi$ by \myref{Eq.~(\ref{eq:outer-opt})}.}
\label{fig:main}
\end{figure*}

\subsection{Framework Formulation}
In this study, we focus on mitigating the issue of catastrophic forgetting of rehearsal-based models with an online CL scenario. As previously mentioned, existing rehearsal-based models often struggle with catastrophic distribution changes caused by dynamic data streams over time. To tackle this challenge, unlike prior methods that regard shifts in label or feature distributions, we suggest directly modeling the catastrophic distribution change for posterior probability $\mathbb P(Y|X)$, enabling the original classification model $f_\theta$ to learn a stable knowledge consolidation for all previous tasks.

The main methodology involves the design of a Continual Bias Adaptor (CBA) denoted as $g_\phi$, which serves two key purposes: 1) Dynamically augmenting the classification network $f_\theta$ to produce more diverse posterior distributions by adjusting the parameters of $g_\phi$ (where $\phi$ can be regarded as hyper-parameters), aiming to address the catastrophic posterior change. 2) Guiding the original classification model $f_\theta$ to fit an implicit posterior that tends to achieve a stable consolidation of knowledge from previously learned tasks. In summary, during the training \wqza{stage}, for a given training example $x^{trn}$, its posterior probability is modified by the augmented classification network $\mathcal F_{\theta, \phi}=g_\phi \circ f_\theta$ in an online CL manner, which can be formulated as
\begin{equation}
    \tilde{y}^{trn} = \mathcal F_{\theta, \phi}(x^{trn})= g_{\phi} \circ f_{\theta}(x^{trn}),
\label{eq:ba_output}
\end{equation}
where $\circ$ is the function composition operator and $g_\phi$ is a lightweight network. 
The augmented classification network $\mathcal F_{\theta, \phi}$ is firstly updated to learn new knowledge from the training data $\mathcal B^{trn}=\mathcal B^t \cup \mathcal B^{buf}$ that minimizes the rehearsal-based empirical risk, \textit{i.e.},
\begin{equation}
    \theta^*(\phi)=\arg\min_\theta \mathcal{L}^{trn} \left(\mathcal{B}^{trn}; \mathcal F_{\theta, \phi} \right),
\label{eq:inner-loss}
\end{equation}
where $\phi$ is a hyper-parameter of the optimal $\theta^*$. Note that different rehearsal-based loss functions can be used as the training loss $\mathcal{L}^{trn}$. Here we only take ER as an example for simplicity and more details can be found in \myref{Appendix~D}.

The ultimate objective of our method is to protect the original classification network $f_\theta$ from catastrophic distribution shift while achieving a stable knowledge consolidation across different tasks. To this end, we further keep tracking the performance of the classification network to prevent catastrophic forgetting, which requires that $f_{\theta^*(\phi)}$, obtained by minimizing the rehearsal-based empirical risk \myref{Eq.~(\ref{eq:inner-loss})}, maximizes the performance of all previously seen data. However, accessing all of this historical data is unfeasible in CL, we approximate it by the empirical risk over the memory buffer data, \textit{i.e.},
\begin{equation}
\begin{split}
    \phi^* &= \arg\min_\phi \mathcal{L}^{buf} \left(\mathcal{B}^{buf}; f_{\theta^*(\phi)} \right), \\
    &= \arg\min_\phi \frac{1}{|\mathcal{B}^{buf}|} \sum_{x,y \in \mathcal{B}^{buf}} L\left(f_{\theta^*(\phi)}(x), y\right).
\end{split}
\label{eq:outer-loss}
\end{equation}
This objective function aims to find the optimal CBA such that the optimized classification network performs well on the memory buffer data, which acts as a stable consolidation of knowledge from the learned tasks.

Indeed, \myref{Eq.~(\ref{eq:inner-loss})} and \myref{Eq.~(\ref{eq:outer-loss})} formulate a bi-level learning framework and the main flowchart is illustrated in \myref{Fig.~\ref{fig:main}}. In the inner loop \myref{Eq.~(\ref{eq:inner-loss})}, the classification network is updated to learn new knowledge and rehearse old knowledge from $\mathcal B^{trn}$ with the help of the CBA module. In the outer loop \myref{Eq.~(\ref{eq:outer-loss})}, the CBA $g_\phi$ are updated from $\mathcal B^{buf}$ to consolidate the previously learned knowledge against catastrophic posterior change. In subsequent sections, we elaborate on the proposed CBA module and bi-level optimization algorithm in detail:

\medskip
\noindent\textbf{The Learning of CBA.}
For the bi-level optimization framework \myref{Eq.~(\ref{eq:inner-loss}) and (\ref{eq:outer-loss})}, there is no closed-form solution, especially in the deep learning field~\cite{bilevel}. That is because the optimum of $\theta$ and $\phi$ are nested with each other. Therefore, we approximately update $\theta$ and $\phi$ using a gradient-based optimization method following~\cite{learn2reweight, meta-weight-net}.

(1) \textbf{Update $\theta$.}
Given the CBA parameter $\phi^k$ at iteration step $k$, we update the classification network parameter $\theta^k$ in \myref{Eq.~(\ref{eq:inner-loss})} by one-step stochastic gradient descent (SGD):
\begin{equation}
    \theta^{k+1}(\phi^k) = 
    \theta^{k} - \alpha \cdot \nabla_{\theta} \mathcal{L}^{trn} \left(\mathcal{B}^{trn}; \mathcal F_{\theta^k, \phi^k} \right),
\label{eq:inner-opt}
\end{equation}
where $\alpha$ is the inner-loop learning rate.

(2) \textbf{Update $\phi$.}
With the one-step updated $\theta^{k+1}$, a function of $\phi$, we further optimize $\phi$ in \myref{Eq.~(\ref{eq:outer-loss})} as follows:
\begin{equation}
    \phi^{k+1} = 
    \phi^{k} - \beta \cdot \nabla_{\phi} \mathcal{L}^{buf} \left( \mathcal{B}^{buf}; f_{\theta^{k+1}(\phi)} \right),
\label{eq:outer-opt}
\end{equation}
where $\beta$ is the outer-loop learning rate.
Note that $\nabla_{\phi} \mathcal{L}^{buf}$ in \myref{Eq.~(\ref{eq:outer-opt})} involves a second-order derivative, which can be easily implemented by automatic differentiation systems such as Pytorch~\cite{pytorch}. The detailed calculations are provided in \myref{Appendix~A}.
Additionally, to alleviate the calculation burden of this second-order derivation, we assume that $\phi$ only depends on the last linear layer of the classification network. Consequently, we only need to unroll the second-order derivation of the linear classification layer in \myref{Eq.~(\ref{eq:outer-opt})}. As the linear layer typically comprises a small number of parameters compared to the entire classification network, our proposed algorithm is more efficient than other bi-level optimization algorithms \cite{MAML, DARTS, learn2reweight, meta-weight-net}.
Please refer to \myref{Appendix~H} for a comprehensive discussion on computation and GPU memory utilization.
The complete training process is detailed in \myref{Alg.~\ref{alg:cba-training}}.

\begin{figure}[t]
\begin{center}
   \includegraphics[width=0.8\columnwidth]{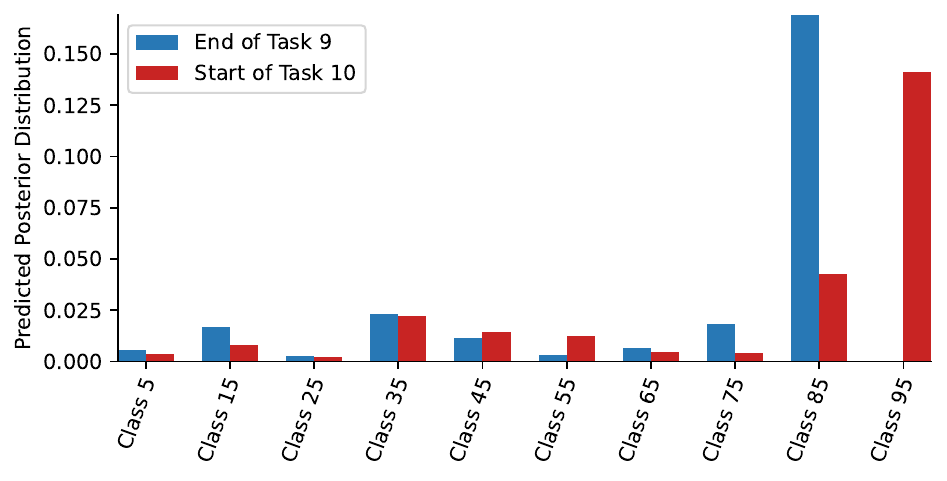}
\end{center}
\vspace{-5mm}
\caption{Illustration of the posterior distribution $\hat{y}$ predicted by the classification network before and after the task transition timestamp. We train the classification network with class-specific CBA on Split CIFAR-100 and take test samples of the 85th class (which belongs to the 9th task) as an example. The posterior distribution $\hat{y}$ contains the probabilities for all 100 classes and we only show that of one class from each task for clarity. Before the task transition (end of the 9th task), the classification network assigns a high posterior probability to the 85th class. However, after the task transition (start of the 10th task), the posterior probability of the 85th class drops dramatically, seriously biased toward new tasks.} 
\label{fig:distrib_change_cls}
\end{figure}

\subsection{Design of Dual-CBA}
\label{subsec:design_cba}
In this subsection, we delve into the architectural design of the CBA module $g_\phi$, which plays a vital role in adapting to catastrophic posterior changes in continual learning.

\medskip
\noindent\textbf{Class-specific CBA.}
An intuitive design for the CBA module is to element-wisely adjust the posterior distribution $\hat{y} \in \mathbb{R}^{\left|\mathcal{C}_t\right|}$ output by the original classification network $f_\theta$, where $\mathcal{C}_t$ denotes the set of all seen classes. This adjustment essentially defines a mapping from $\hat{y}$ to a corresponding adapted posterior distribution $\tilde{y} \in \mathbb{R}^{\left|\mathcal{C}_t\right|}$. To achieve this, we use a multi-layer perceptron (MLP) network $g_\omega^{spc}$ parameterized by $\omega$ as a part of the CBA module, which can be formulated as:
\begin{equation}
    \tilde{y}^{spc} = g_\omega^{spc}(\hat{y}): \mathbb{R}^{\left|\mathcal{C}_t\right|} \rightarrow \mathbb{R}^{\left|\mathcal{C}_t\right|},
\end{equation}
where $\hat{y} = \left[ p\big(y=1\big|x\big), p\big(y=2\big|x\big), \cdots, p\big(y=\left|\mathcal{C}_t\right| \big| x \big) \right]^T$ and $\tilde{y}^{spc} = \left[ p\big(y=1\big|\hat{y}\big), p\big(y=2\big|\hat{y}\big), \cdots, p\big(y=\left|\mathcal{C}_t\right| \big| \hat{y} \big) \right]^T$. Such a class-specific CBA is known as a universal approximator, capable of fitting almost any continuous function~\cite{MLP-universal-approx} and thus can adapt to various posterior distribution changes. Importantly, our conference version~\cite{CBA} has validated the effectiveness of the class-specific CBA on assimilating task-recency bias upon the rehearsal-based CL methods.

\medskip
\noindent\textbf{Class-agnostic CBA.}
Albeit fast adaptation to drastic posterior distribution changes, the proposed class-specific CBA module suffers from a substantial stability gap, where performance significantly drops upon starting to learn new tasks, followed by a fast recovery phase. This to some extent does not meet the need for online CL to conduct performance evaluations at any time. Thus, a natural question is: \textit{what causes the class-specific CBA to aggregate the stability gap problem, and how can this challenge be effectively addressed?}

The primary reason is that the class-specific CBA cannot adapt to the abrupt change in the posterior distribution upon starting to learn a new task. 
To investigate this, we visualize the difference between the posterior probability $\hat{y}$ predicted by the classification network before and after the task transition timestamp in \myref{Fig.~\ref{fig:distrib_change_cls}}. We can see that the classification network $f_\theta$ produces a relatively high posterior probability for the 85th class, a new class introduced in task 9. However, when task 10 comes in, the probability of the 85th class is significantly lower than before. Since the posterior probability predicted by the classification network is the input of the class-specific CBA, the corruption of these inputs for task 9 hinders the class-specific CBA from appropriately adjusting its posterior probabilities.

The second reason is that the class-specific CBA cannot immediately adjust the posterior probability of the classes in the new task.
When a new task comes in, the class-specific CBA fails to accurately accommodate the posterior probability of the new task because it has not yet encountered the incoming new classes. As shown in \myref{Fig.~\ref{fig:distrib_change_cls}}, the arrival of task 10 leads to a high posterior probability for its own classes, such as the 95th class. This indicates that the class-specific CBA fails to control the increasing probability of the new task.

\begin{figure}[t]
\centering
\subfloat[]{
    \includegraphics[width=0.75\columnwidth]{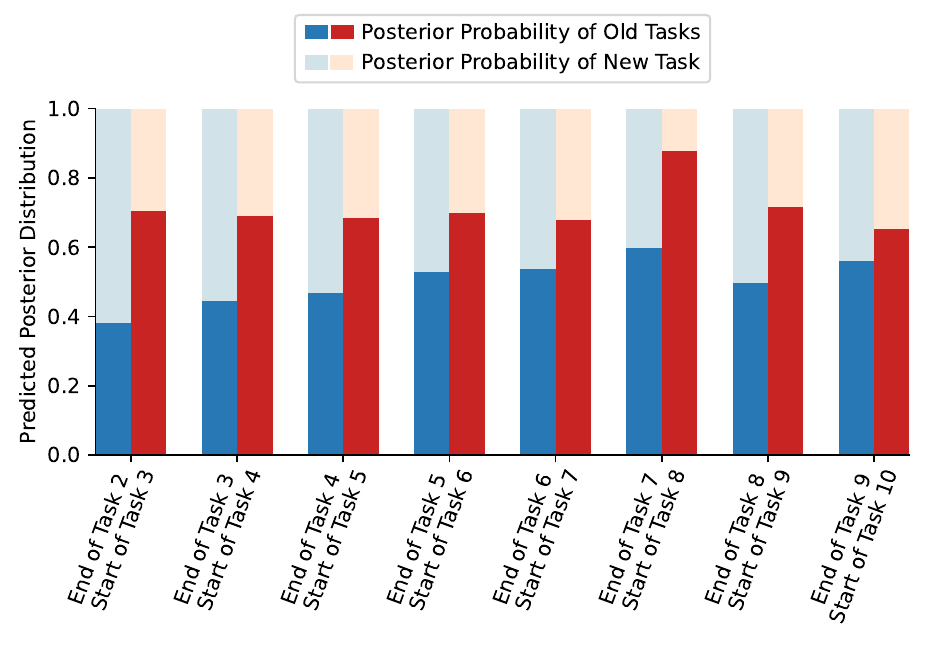}%
    \label{subfig:distrib_change_tsk_old}}
\hspace{8pt}
\subfloat[]{
    \includegraphics[width=0.75\columnwidth]{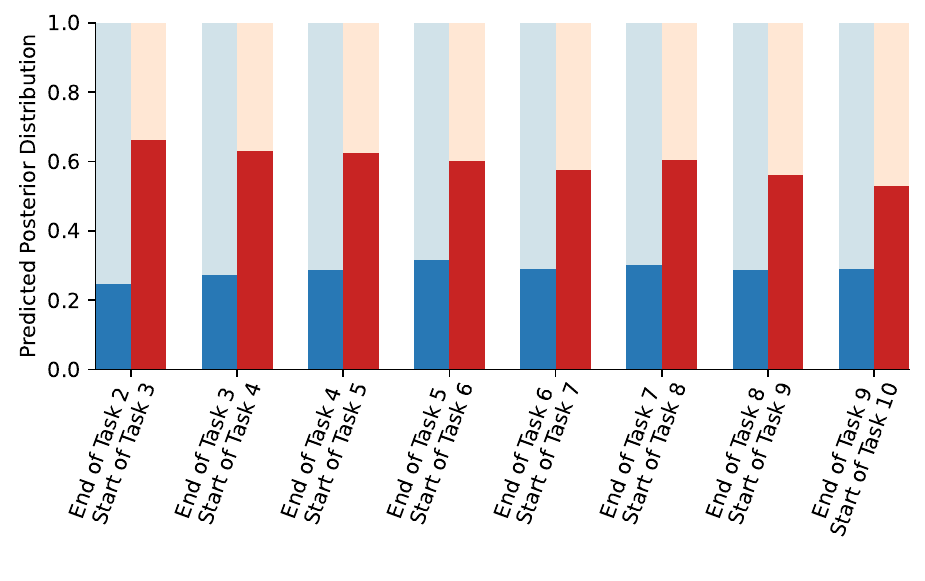}%
    \label{subfig:distrib_change_tsk_new}}
\caption{Illustration of the old-task posterior probability $\hat{y}_{old}$ and the new-task posterior probability $\hat{y}_{new}$ predicted by the classification network at each task transition timestamp during the whole continual learning process: (a) test samples from all old tasks; (b) test samples from the new task. The classification network is trained by ER on Split CIFAR-100, where the relationship between the old- and new-task probabilities shows a stable trend as training progresses.}
\vspace{-3mm}
\label{fig:distrib_change_tsk}
\end{figure}

To address the aforementioned challenges, instead of directly adjusting the posterior probability in a class-specific manner, we resort to modeling a stable relationship between new and old tasks to adapt to the posterior distribution shift while avoiding a significant stability gap upon starting to learn new tasks. Let $\mathcal{C}_{new}$ and $\mathcal{C}_{old}$ denote the class sets of the new task and old tasks, respectively. Our main idea is to adjust the posterior probability of the new task $\hat{y}_{new} = \sum_{c \in \mathcal{C}_{new}} p(y=c|x)$ and that of all the old tasks $\hat{y}_{old} = \sum_{c \in \mathcal{C}_{old}} p(y=c|x)$, given their relatively stable numerical relationship throughout the continual learning \wqza{process} as shown in \myref{Fig.~\ref{fig:distrib_change_tsk}}.
To achieve this, we design a novel class-agnostic CBA $g_\nu$ parameterized by $\nu$ as follows:
\begin{equation}
    (\tilde{y}_{old}, \tilde{y}_{new}) = g_\nu^{agn}(\hat{y}_{old}, \hat{y}_{new}): \mathbb{R}^2 \rightarrow \mathbb{R}^2.
\end{equation}
Note that the class-agnostic CBA treats the posterior probabilities of the classes in the new task as a whole, and does the same for the old task. Then, for a specific class, the adjustment of its posterior probability is formulated by averaging $\tilde{y}_{old}$ or $\tilde{y}_{new}$, and it can be formulated as 
\begin{equation}
    p(y=c|\hat{y}) = 
    \begin{cases} 
        \frac{1}{\left|\mathcal{C}_{old}\right|} \tilde{y}_{old} & c \in \mathcal{C}_{old} \\
        \frac{1}{\left|\mathcal{C}_{new}\right|} \tilde{y}_{new} & c \in \mathcal{C}_{new}
    \end{cases}
\end{equation}
and the final adapted posterior probabilities are given by $\tilde{y}^{agn} = \left[ p\big(y=1\big|\hat{y}\big), p\big(y=2\big|\hat{y}\big), \cdots, p\big(y=\left|\mathcal{C}_t\right| \big|\hat{y}\big) \right]^T$.

The proposed class-agnostic CBA module serves as a robust and transferable bias adaptor, effectively adapting to sudden changes in posterior distributions after training on a few pairs of new and old tasks. Specifically, we can analyze its adaptation mechanisms for both the preceding old task and the incoming new task. We assume that the task transition timestamp is $t+1$, and we have
\begin{itemize}
    \item For the incoming $(t+1)$-th new task, the class-agnostic CBA module has processed $t$ pairs of old and new tasks, and its current weights can produce a valuable initialization or experience learned from previous data. Therefore, the class-agnostic CBA can quickly adapt to new tasks.
    \item For the $t$-th task, the class-agnostic CBA module naturally treats it as an old task at the $t+1$ timestamp. Similarly, CBA leverages the experience learned from previous old tasks to adjust its posterior distribution. This robust initialization makes class-agnostic CBA effectively adapt to the changed posterior distribution.
\end{itemize}
In summary, the proposed class-agnostic CBA can produce a robust initialization for both the $(t+1)$-th new task and the $t$-th old task, ensuring to adapt to the suddenly changed posterior distribution quickly. Consequently, the class-agnostic CBA can help the classification model learn the stable posterior distribution and mitigate the stability gap immediately upon starting to learn a new task.

\medskip
\noindent\textbf{Dual-CBA.}
For the class-agnostic CBA, we treat all classes equally. However, it is essential to model the discrepancy between classes in the CL process. 
To capture both the stable relationship between new and old tasks and the refined class-wise relationship, we integrate the class-agnostic CBA $g^{agn}_\nu$ and the class-specific CBA $g^{spc}_\omega$ into a unified module $g_\phi$ (referred to as Dual-CBA) with $\phi = \{\omega, \nu\}$ denoting its parameters.
Specifically, we average $\tilde{y}^{spc}$ and $\tilde{y}^{agn}$ as the final adapted posterior distribution $\tilde{y}$, that is,
\begin{equation}
    \tilde{y} = g_{\phi}(\hat{y}) := \frac{1}{2} (\tilde{y}^{spc} + \tilde{y}^{agn}).
\end{equation}
Since the class-specific CBA may aggregate the stability gap as aforementioned, we reinitialize the class-specific CBA modules to avoid incorrect adjustments when a new task arrives. This enables our method to be inferred at any time in the learning process of online CL~\cite{ACC_AUC} without being disturbed by the stability gap problem, which will be validated through extensive experiments in \myref{Sec.~\ref{subsec:discussion}} and \myref{Appendix~F}.

Note that the proposed Dual-CBA module is only used in the training stage. In the test stage, the test sample $x^{tst}$ is predicted by $f_\theta$, that is $\hat{y}^{tst} = f_\theta(x^{tst})$. This indicates that our method does not introduce any calculation overhead in the test stage.

\begin{algorithm}[t]
\caption{Training of Dual-CBA in Online CL} 
\label{alg:cba-training}

    \textbf{Input:} new incoming sample batch $\mathcal{B}^t$, memory buffer $\mathcal{M}$ \\
    \textbf{Output:} optimized classification network parameter $\theta^*$ and Dual-CBA parameter $\phi^* = \{\omega^*, \nu^*\}$
    \begin{algorithmic}[1]
        \State Initialize all network parameters $\theta^0, \phi^0=\{\omega^0, \nu^0\}$.
        \State Stop the EMA update of BN population statistics in the classification network.
        \State \textbf{While} $\mathcal{B}^t \neq \emptyset$ \textbf{do}
            \State \qquad \textbf{If} new task \wqza{arrives}:
                \State \qquad \qquad Initialize the class-specific CBA parameter $\omega$.
            \Statex \qquad \textcolor{gray}{\# Inner-loop optimization:}
            \State \qquad $\mathcal{B}^{trn} = \mathcal{B}^t \cup \mathcal{B}^{buf}, \mathcal{B}^{buf} \subset \mathcal{M}$  \textcolor{gray}{\# Inner training data}
            \State \qquad Compute the inner-loop loss $\mathcal{L}^{trn}$ by \myref{Eq.~(\ref{eq:inner-loss})}.
            \State \qquad Update classifier network parameters $\theta$ by \myref{Eq.~(\ref{eq:inner-opt})}.
            \Statex \qquad \textcolor{gray}{\# Outer-loop optimization:}
            \State \qquad $\mathcal{B}^{buf} \subset \mathcal{M}$  \textcolor{gray}{\# Outer training data}
            \State \qquad Compute the outer-loop loss $\mathcal{L}^{buf}$ by \myref{Eq.~(\ref{eq:outer-loss})}.
            \State \qquad Update CBA parameters $\phi = \{\omega, \nu\}$ by \myref{Eq.~(\ref{eq:outer-opt})}.
    \end{algorithmic} 
\end{algorithm}

\begin{algorithm}[t]
\caption{Testing of Dual-CBA in Online CL} 
\label{alg:cba-testing}

    \textbf{Input:} testing dataset $\mathcal{D}^{test} = \{ x^{test}_i \}_{i=1}^{N}$, memory buffer $\mathcal{M}$, optimized classification network parameter $\theta^*$ \\
    \textbf{Output:} predicted results $\{\hat{y}^{test}_i\}_{i=1}^{N}$
    \begin{algorithmic}[1]
        \State Using $\mathcal{M}$ to update the BN population statistics of the classification network by EMA.
        \State Predict the results by $\hat{y}^{test}_i = f_{\theta^*}(x^{test}_i)$ for any $i=1, \cdots, N$.
    \end{algorithmic} 
\end{algorithm}

\subsection{Incremental Batch Normalization}
\label{subsec:IBN}
In this subsection, we further address the feature bias introduced during the training stage by the commonly used BN~\cite{BN} of our proposed bi-level framework. Specifically, BN calculates batch statistics of each feature map for normalization during training, while using population statistics estimated by an exponential moving average (EMA) during testing. In our proposed bi-level optimization framework in the conference version, the population statistics of BN are only updated in the inner loop together with the classification network parameters $\theta$. Unfortunately, these population statistics estimated by EMA in the inner loop are seriously biased toward the current task because the training data are dominated by new tasks, leading to a biased feature extraction during testing. For simplicity, we take the population mean used for testing at task $t$ as an example. Specifically, we denote it as $\mu_{test}^t$, and it is updated using an EMA scheme as follows:
\begin{equation}
    \mu^{t,k}_{test} = (1 - \eta) \cdot \mu^{t,k-1}_{test} + \eta \cdot \mu^{t,k}_{train},
\end{equation}
where $\mu^{t,k}_{train}$ is the feature mean of the $k$-th batch of training data $\mathcal{B}^{trn}$ at task $t$. Then, we have
\begin{equation}
    \mu^{t,k}_{test} = (1 - \eta)^k \cdot \mu^{t,0}_{test} + \sum^k_{i=1} (1 - \eta)^{k - i} \eta \cdot \mu^{t,i}_{train},
\end{equation}
where $\mu^{t,0}_{test}$ is initialized as the population mean calculated at the final of the last task $t-1$. It can be observed that $\mu^{t,0}_{test}$ encodes the feature mean of previous tasks and the corresponding weight is exponentially decreasing as current task training progresses. Additionally, the batch mean $\mu^{t,k}_{train}$ is dominated by the current new task data because the number of the memory buffer samples is much smaller than that of incoming new samples. Consequently, the population mean updated by EMA in the inner loop is seriously biased towards new tasks, where the population variance has the same tendency.

Note that this feature bias is not reflected in the posterior distribution during training, where the proposed Dual-CBA module cannot correct the biased feature. To achieve unbiased estimations of these population statistics, we should estimate them on the seen data from all tasks, which is unrealistic in continual learning. Therefore, from the bi-level optimization perspective, an intuitive idea is to update these population statistics by the balanced buffer data set used in the outer loop. Intrinsically, the memory buffer obtained via reservoir sampling is an excellent approximation of the entire data distribution for all seen tasks in an ideal case. Based on this principle, we propose a straightforward yet effective method to provide approximately unbiased population statistics in continual learning, termed Incremental Batch Normalization (IBN). Specifically, we stop the EMA updating of BN statistics in the inner loop optimization and estimate unbiased population statistics using the memory buffer in the outer loop optimization. For the implementation of IBN, we further refine the process by replacing the update of population statistics during the training stage with estimations based on balanced buffer data before the testing stage. 

This procedure does not interfere with the optimization of our bi-level learning framework, which can effectively assimilate the feature discrepancy of the classification model during inference by estimating more accurate population statistics in BN for all seen tasks.
Specifically, the population statistics estimated by IBN can be formulated as follows:
\begin{equation}
    \mu^{t}_{test} = \sum^K_{i=1} (1 - \eta)^{K - i} \eta \cdot \mu^{t,i}_{buf},
\end{equation}
where $\mu^{t,i}_{buf}$ is the feature mean of the $k$-th batch of memory buffer samples $\mathcal{B}^{t} \in \mathcal{M}$ at task $t$, and $K$ is the total number of moving average.
The implementation of IBN is simple as it only requires forward propagation of the small balanced buffer data without any back-propagation, and the population statistics can be updated by EMA automatically, which does not introduce too much calculation burden.

\section{Theoretical Analysis} 
\label{sec:theory}
In this section, we provide a theoretical analysis of our method. Firstly, we explain how our bi-level optimization method effectively prevents forgetting from the perspective of gradient alignment. Then we illustrate the general intuition by a linear formulation of the bi-level optimization framework.

\subsection{Gradient Alignment in Dual-CBA}
\label{subsec:gradient_align}
The following theorem reveals that the proposed bi-level optimization inherently establishes gradient alignment between the loss on the training set $\mathcal{B}^{trn}$ and the memory buffer $\mathcal{B}^{buf}$.

\medskip
\begin{theorem}
\textit{
Let $\mathcal{G}^{buf} \triangleq \frac{\partial \mathcal{L}^{buf}\left(\mathcal{B}^{buf}; f_{\theta^k(\phi)}\right)}{\partial \theta^k}$ and $\mathcal{G}^{trn} \triangleq \frac{\partial \mathcal{L}^{trn}\left(\mathcal{B}^{trn}; \mathcal{F}_{\theta^k, \phi}\right)}{\partial \theta^k}$ denote the gradients of the outer-loop and inner-loop losses with respect to the classification model parameter $\theta$, respectively. If the outer-loop loss $\mathcal{L}^{buf}(\cdot; f_\theta)$ is $\eta$ gradient Lipschitz continuous, then the bi-level optimization \myref{Eq.~(\ref{eq:inner-loss}) and (\ref{eq:outer-loss})} potentially guarantees an alignment between $\mathcal{G}^{buf}$ and $\mathcal{G}^{trn}$, that is}
\begin{equation}
    \left\langle \mathcal{G}^{buf}, \mathcal{G}^{trn} \right\rangle 
    \geq
    \frac{\alpha \eta}{2}
    \left\| \mathcal{G}^{trn} \right\|_2^2,
\label{eq:gradient_alignment}
\end{equation}
\textit{where $\alpha > 0$ is the inner-loop learning rate and $\eta > 0$ is the Lipschitz constant.}
\label{thm:gradient_alignment}
\end{theorem}

\medskip
The detailed proof of \myref{Theorem~\ref{thm:gradient_alignment}} is shown in \myref{Appendix~B}.
This theorem demonstrates that optimizing our proposed bi-level framework \myref{Eq.~(\ref{eq:inner-loss}) and (\ref{eq:outer-loss})} encourages the angle between the gradients $\mathcal G ^{buf}$ and $\mathcal G ^{trn}$ to be as small as possible, which reveals two insights into our algorithm. On the one hand, this theorem indicates that our approach will push the gradients of the classification network on the training data (with the Dual-CBA module) to those on the buffer data (without the Dual-CBA module). This property potentially encourages the Dual-CBA module to mitigate the training bias present in the training data. That is why our algorithm can effectively mitigate the task-recency bias in the CL process. On the other hand, this theorem shows an alignment between the gradients of the classification network on the training data and buffer data, which regularizes the updating of the classification network on the new task and ensures it does not deviate from the previous updating direction too much. This gradient alignment can effectively alleviate forgetting in CL, which has been demonstrated in many previous gradient-alignment-based methods~\cite{AGEM, GEM}. However, these works calculate the classification network gradient on training data without accounting for task-recency bias, where the network may seriously bias to new tasks, resulting in an imprecise gradient alignment. In contrast, our proposed Dual-CBA method mitigates this bias, ensuring more accurate gradient alignment and improved performance.

\subsection{Closed-form Solution of Linear Dual-CBA}
\label{subsec:close-form}
We herein consider a convex model to delve into the insight of our proposed Dual-CBA model. Specifically, we reformulate Dual-CBA as a linear model where the origin and augmented classification networks are represented as:
\begin{equation}
\begin{split}
    {\rm \textbf{Original:}} \quad \hat{Y}_t &= (X_t)^T \theta^*_t + e_t \\
    {\rm \textbf{Augmented:}} \quad \tilde{Y}_t &= \hat{Y}_t \phi^* + e_t, \\
\end{split}
\label{eq:linear-forward}
\end{equation}
where $X_t \in \mathbb{R}^{p \times n_t}$ is the input data and $\hat{Y}_t, \tilde{Y}_t \in \mathbb{R}^{n_t \times c_t}$ denote outputs of the original and augmented classification networks, respectively. The parameters of the classification model are $\theta^*_t \in \mathbb{R}^{p \times c_t}$ and the Dual-CBA module is represented as $\phi^* \in \mathbb{R}^{c_t \times c_t}$, where we omit the notation of $\nu$ for clarity. Additionally, we assume the Gaussian feature and noise of this linear model following~\cite{theory1} with $e_t \sim \mathcal{N}(0, \sigma^2_e)$ denoting the noise vector.

We consider the mean square error (MSE) loss as the convex objective function. In this case, the proposed bi-level optimization framework can be represented as follows:
\begin{equation}
\begin{split}
    &\phi^*_t = \arg\min_\phi \frac{1}{n^{buf}_t} \lVert (X^{buf}_t)^T \theta^*_t(\phi) - Y^{buf}_t \rVert^2_2 \\
    {\rm s.t.} \quad 
    &\theta^*_t(\phi^*_t) = \arg\min_\theta \frac{1}{n^{trn}_t} \lVert (X^{trn}_t)^T \theta_t\phi^*_t - Y^{trn}_t \rVert^2_2. \\
\end{split}
\label{eq:linear-model}
\end{equation}
Note that $\theta_t \in \mathbb{R}^{p \times c_t}$ is the weight matrix and $\theta^*_t(\phi)$ represents a function of $\phi  \in \mathbb{R}^{c_t \times c_t}$.
Thanks to the excellent properties of convex optimization, we can get the closed-form solution of this bi-level optimization framework, which can be summarized as follows:

\medskip
\begin{theorem}
    \textit{Let $(X)^\dagger = (XX^T)^{-1} X$ and $A = (X^{buf}_t)^T (X^{trn}_t)^\dagger = (X^{buf}_t)^T \left( X^{trn}_t (X^{trn}_t)^T \right)^{-1} X^{trn}_t$. If the parameter of CBA $\phi$ is not singular, then the closed-form solution of the linear bi-level optimization \myref{Eq.~(\ref{eq:linear-model})} can be represented as}
    \begin{equation}
    \begin{split}
        \theta^*_t &= (X^{trn}_t)^\dagger Y^{trn}_t (\phi^*_t)^{-1}, \\
        \phi^*_t &= 
    \left[ (Y^{trn}_t)^T A^T \textcolor{red}{Y^{buf}_t} \right]^{-1}
    \left[ (Y^{trn}_t)^T A^T \textcolor{red}{A Y^{trn}_t} \right].
    \end{split}
    \label{eq:linear-solution}
    \end{equation}
    \label{thm:linear-solution}
\end{theorem}

The proof of this theorem can be found in the \myref{Appendix~B}.
It can be observed that the optimal solution of $\phi^*_t$ in \myref{Eq.~(\ref{eq:linear-solution})} actually measures the discrepancy between $\small{Y^{buf}_t}$ and $\small{A Y^{trn}_t = (X^{buf}_t)^T (X^{buf}_t)^\dagger Y^{trn}_t}$. Intuitively, if we multiply these two terms by $(X^{buf}_t)^\dagger$ on the left of the two sides of this equation, we can find that this discrepancy depends on the difference between $(X^{buf}_t)^\dagger Y^{buf}_t$ and $(X^{trn}_t)^\dagger Y^{trn}_t$, which are the optimal solutions of the linear model on the buffer data $\lVert (X^{buf}_t)^T \theta - Y^{buf}_t \rVert^2$ and that on the training data $\lVert (X^{trn}_t)^T \theta - Y^{trn}_t \rVert^2$, respectively. Therefore, the proposed Dual-CBA module $\phi^*_t$ captures the relationship between the optimal linear solutions on training data and buffer data, which can effectively feedback to the optimization of the classification network. Specifically, the optimal $\phi^*_t$ regularizes the optimal solution of $\theta^*_t$ by modifying the optimal linear solution on the training data $(X^{trn}_t)^\dagger Y^{trn}_t$, ensuring the classification network pay more attention to the balanced buffer data $\{X^{buf}_t, Y^{buf}_t\}$ rather than only overfit on the current training data $\{X^{trn}_t, Y^{trn}_t\}$.

\section{Experiments}
\label{sec:experiments}
To validate the effectiveness of the proposed method, we compare our Dual-CBA to multiple approaches on various datasets under online CL, semi-supervised CL, blurry tasks, and offline CL settings. We also conduct extensive ablation experiments to analyze different components of our approach.

\subsection{Experiment Settings}
\label{subsec:exp-settings}

\medskip
\noindent\textbf{Experimental datasets.}
Following \cite{DER}, we experiment with three widely-used datasets: \textbf{Split CIFAR-10}~\cite{cifar}, \textbf{Split CIFAR-100} and \textbf{Split Tiny-ImageNet}~\cite{tinyimg}. Concretely, Split CIFAR-10 contains five binary classification tasks, which are constructed by evenly splitting ten classes of CIFAR-10. Split CIFAR-100 and Split Tiny-ImageNet both have longer task sequences with each comprising ten disjoint tasks. Specifically, Split CIFAR-100 includes ten tasks with 10 classes each, while Split Tiny-ImageNet includes ten tasks with 20 classes each (see \myref{Appendix~D} for details of the three datasets).

\medskip
\noindent\textbf{Evaluation metrics.} To comprehensively evaluate all comparison methods, we consider the following metrics:
\begin{itemize} 
    \item \textbf{Average Accuracy} (ACC $\uparrow$): This metric calculates the average accuracy of the model trained on all tasks, \textit{i.e.} ${\rm{ACC}} = 1/T \sum^{T}_{t=1} a_{t, T}$, where $a_{i, j}$ represents the accuracy of the task $i$ after training on the task $j$. $\uparrow$ indicates that a higher ACC value corresponds to better performance.
    
    \item \textbf{Forgetting Measure} (FM $\downarrow$): This metric averages the differences between the best accuracy and the final accuracy, \textit{i.e.} ${\rm{FM}} = 1/T \sum^{T}_{t=1} a^*_t - a_{t, T}$, where the $a^*_t$ is the best accuracy of task $t$ in the whole training process. $\downarrow$ indicates that a lower FM value corresponds to better performance.
    
    \item \textbf{Area Under the Curve of Accuracy} (${\rm{{ACC}_{AUC}}} \uparrow$): This metric is the area under the curve of the accuracy~\cite{ACC_AUC}, \textit{i.e.,} ${\rm{{ACC}_{AUC}}} = \sum_i \bar{a}(i\cdot \triangle n)\cdot \triangle n$, where $\bar{a}(i)$ represent the average accuracy when the model training at step $i$, and $\triangle n$ is the interval training step which is 5 for faster evaluation in our experiments. $\uparrow$ indicates that a higher ${\rm{{ACC}_{AUC}}}$ value corresponds to better performance.
\end{itemize}

\medskip
\noindent\textbf{Implementation details.} We adopt the commonly used ResNet-18~\cite{resnet} as our backbone~\cite{CLSER, DER, bic}, and train all methods using the Stochastic Gradient Descent (SGD) optimizer. We use Adam~\cite{adam} to optimize the proposed Dual-CBA module and set the learning rate as 0.001 for Split CIFAR-10, and 0.01 for Split CIFAR-100 and Split Tiny-ImageNet. To reduce the variability in experimental results, each reported result in the online CL setting is averaged over 10 repeated runs, and each result in the offline CL is averaged over 5 runs. More details about the baselines and implementations are listed in \myref{Appendix~D}.

\begin{table*}[]
\caption{Main results (ACC, higher is better, and FM, lower is better) on the three datasets with different memory buffer sizes. Our method applied on 4 baselines is shown with gray cells. \textbf{Bold} means the best results in all comparison methods and the gains of Dual-CBA comparing the corresponding baselines are shown in \textcolor{red}{red} color.}
\label{tab:main-results}
\centering
\begin{tabular}{ccccccccccccc}
\hline
                             & \multicolumn{4}{c}{Split CIFAR-10}                                                                        & \multicolumn{4}{c}{Split CIFAR-100}                                                                       & \multicolumn{4}{c}{Split Tiny-ImageNet}                                                                  \\ \cline{2-13} 
                             & \multicolumn{2}{c}{M=0.2k}                          & \multicolumn{2}{c}{M=0.5k}                          & \multicolumn{2}{c}{M=2k}                            & \multicolumn{2}{c}{M=5k}                            & \multicolumn{2}{c}{M=2k}                            & \multicolumn{2}{c}{M=5k}                           \\ \cline{2-13} 
\multirow{-3}{*}{Method}     & ACC $\uparrow$           & FM $\downarrow$          & ACC $\uparrow$           & FM $\downarrow$          & ACC $\uparrow$   & FM $\downarrow$          & ACC $\uparrow$           & FM $\downarrow$          & ACC $\uparrow$           & FM $\downarrow$          & ACC $\uparrow$           & FM $\downarrow$         \\ \hline
iCaRL~\cite{icarl}           & 40.99                    & 26.84                    & 44.5                     & 24.87                    & 9.13                     & 7.79                     & 9.13                     & 8.14                     & 4.03                     & 4.93                     & 4.03                     & 5.15                    \\
LUCIR~\cite{LUCIR}           & 23.59                    & 35.59                    & 24.63                    & 31.89                    & 8.28                     & 16.07                    & 12.31                    & 14.02                    & 4.47                     & 20.4                     & 5.29                     & 20.28                   \\
BiC~\cite{bic}               & 27.71                    & 66.45                    & 35.47                    & 47.92                    & 16.32                    & 36.7                     & 20.89                    & 32.33                    & 5.43                     & 40.14                    & 7.5                      & 38.52                   \\
ER-ACE~\cite{online-er-ace}  & 41.49                    & 20.84                    & 46.35                    & 18.98                    & 24.95                    & 7.67                     & 26.54                    & 7.25                     & 17.89                    & 7.04                     & 19.04                    & 6.9                     \\
SS-IL~\cite{SSIL}            & 37.92                    & 15.64                    & 41.22                    & 11.46                    & 24.9                     & 9.85                     & 25.6                     & 10.23                    & 17.91                    & 7.93                     & 18.53                    & 8.26                    \\ \hline
ER                           & 36.36                    & 49.81                    & 44.16                    & 38.13                    & 21.66                    & 33.21                    & 23.38                    & 33.19                    & 15.03                    & 31.48                    & 16.99                    & 30.48                   \\
ER-CBA                       & 38.13                    & 43.94                    & 45.52                    & 29.96                    & 25.56                    & 9.86                     & 28.55                    & 8.27                     & 17.69                    & 12.85                    & 20.76                    & 10.53                   \\
\rowcolor{gray!40} 
ER-Dual-CBA (ours)               & \textbf{41.40}           & \textbf{28.24}           & \textbf{49.32}           & \textbf{14.33}           & \textbf{29.14}           & \textbf{8.90}            & \textbf{31.08}           & \textbf{7.29}            & \textbf{20.81}           & \textbf{10.55}           & \textbf{23.84}           & \textbf{8.13}           \\
Gains                        & \textcolor{red}{+ 5.04}  & \textcolor{red}{-21.57}  & \textcolor{red}{+ 5.16}  & \textcolor{red}{-23.80}  & \textcolor{red}{+ 7.48}  & \textcolor{red}{-24.31}  & \textcolor{red}{+ 7.70}  & \textcolor{red}{-25.90}  & \textcolor{red}{+ 5.78}  & \textcolor{red}{-20.93}  & \textcolor{red}{+ 6.85}  & \textcolor{red}{-22.35} \\ \hline
DER++~\cite{DER}             & 40.62                    & 42.30                    & 49.78                    & 31.58                    & 17.83                    & 43.91                    & 17.36                    & 44.77                    & 11.50                    & 40.68                    & 11.72                    & 40.90                   \\
DER-CBA                      & 44.05                    & 25.64                    & \textbf{50.97}           & 16.15                    & 25.42                    & 13.08                    & 26.59                    & 13.08                    & 18.81                    & \textbf{11.14}           & 20.13                    & \textbf{11.38}          \\
\rowcolor{gray!40} 
DER-Dual-CBA (ours)              & \textbf{44.29}           & \textbf{20.59}           & 50.12                    & \textbf{9.72}            & \textbf{27.93}           & \textbf{10.24}           & \textbf{29.87}           & \textbf{8.31}            & \textbf{20.73}           & 13.36                    & \textbf{22.52}           & 12.65                   \\
Gains                        & \textcolor{red}{+ 3.67}  & \textcolor{red}{-21.71}  & \textcolor{red}{+ 0.34}  & \textcolor{red}{-21.86}  & \textcolor{red}{+10.10}  & \textcolor{red}{-33.67}  & \textcolor{red}{+12.51}  & \textcolor{red}{-36.46}  & \textcolor{red}{+ 9.23}  & \textcolor{red}{-27.32}  & \textcolor{red}{+10.80}  & \textcolor{red}{-28.25} \\ \hline
CLSER~\cite{CLSER}           & 38.68                    & 44.93                    & 47.90                    & 31.39                    & 21.86                    & 35.22                    & 24.00                    & 34.41                    & 15.81                    & 32.13                    & 17.42                    & 31.64                   \\
CLSER-CBA                    & 43.32                    & 29.26                    & 47.68                    & 24.58                    & 26.32                    & 9.50                     & 28.38                    & 9.23                     & 19.32                    & 11.86                    & 21.16                    & 10.39                   \\
\rowcolor{gray!40} 
CLSER-Dual-CBA (ours)            & \textbf{44.03}           & \textbf{20.40}           & \textbf{49.27}           & \textbf{16.09}           & \textbf{29.83}           & \textbf{8.93}            & \textbf{32.29}           & \textbf{7.30}            & \textbf{21.34}           & \textbf{10.65}           & \textbf{24.37}           & \textbf{8.30}           \\
Gains                        & \textcolor{red}{+ 5.35}  & \textcolor{red}{-24.53}  & \textcolor{red}{+ 1.37}  & \textcolor{red}{-15.30}  & \textcolor{red}{+ 7.97}  & \textcolor{red}{-26.29}  & \textcolor{red}{+ 8.29}  & \textcolor{red}{-27.11}  & \textcolor{red}{+ 5.53}  & \textcolor{red}{-21.48}  & \textcolor{red}{+ 6.95}  & \textcolor{red}{-23.34} \\ \hline
RAR~\cite{RAR}               & 42.52                    & 39.48                    & 47.66                    & 33.86                    & 15.53                    & 45.78                    & 14.79                    & 46.84                    & 10.88                    & 39.21                    & 10.44                    & 40.95                   \\
RAR-CBA                      & 44.22                    & 22.62                    & 47.01                    & 15.95                    & 22.47                    & 14.27                    & 22.85                    & 14.24                    & 17.18                    & \textbf{11.62}           & 17.90                    & \textbf{11.97}          \\
\rowcolor{gray!40} 
RAR-Dual-CBA (ours)              & \textbf{44.88}           & \textbf{13.55}           & \textbf{47.68}           & \textbf{11.14}           & \textbf{26.20}           & \textbf{9.82}            & \textbf{26.38}           & \textbf{9.96}            & \textbf{19.72}           & 12.64                    & \textbf{20.83}           & 12.80                   \\
Gains                        & \textcolor{red}{+ 2.36}  & \textcolor{red}{-25.93}  & \textcolor{red}{+ 0.02}  & \textcolor{red}{-22.72}  & \textcolor{red}{+10.67}  & \textcolor{red}{-35.96}  & \textcolor{red}{+11.59}  & \textcolor{red}{-36.88}  & \textcolor{red}{+ 8.84}  & \textcolor{red}{-26.57}  & \textcolor{red}{+10.39}  & \textcolor{red}{-28.15} \\ \hline
\end{tabular}
\end{table*}

\subsection{Comparison on Disjoint Scenario}
\label{subsec:main-results}
\medskip
\noindent\textbf{Dual-CBA can enhance current rehearsal-based methods.}
We first investigate the performance of the proposed method Dual-CBA across different datasets and various memory buffer sizes. Our proposed Dual-CBA can be easily plugged into multiple rehearsal continual learning baselines. To demonstrate it, we choose four commonly used rehearsal-based methods (\textit{i.e.}, ER, DER, RAR, and CLSER) as baselines in our experiments, and the results are summarized in \myref{Table~\ref{tab:main-results}}. Note that the suffixes `CBA' and `Dual-CBA' represent the class-specific CBA in conference version~\cite{CBA} and the proposed Dual-CBA in this paper applied to the corresponding baselines, respectively. It can be observed that: 1) Our Dual-CBA can consistently improve the ACC of all four baselines and significantly reduce their FM across various settings. This indicates that our method can generalize to multiple rehearsal baselines and help them mitigate the task-recency bias by fitting the implicit posterior distribution during CL training. 2) Dual-CBA further shows significant improvements over the previously proposed CBA in~\cite{CBA} across all four baselines, which highlights the effectiveness of our proposed method. For example, for ER on Split CIFAR-10 with 200 buffer samples, ER-CBA yields an improvement in the ACC of 1.77\% and a reduction in FM of 5.84\%. In contrast, ER-Dual-CBA improves the ACC of ER by about 5.04\% and decreases the FM by about 21.57\%. Additional results and analyses can be found in \myref{Appendix~G}.

\medskip
\noindent\textbf{Analysis of other baselines.}
LUCIR employs a weight normalization prior, and BiC designs an additional linear layer, with both methods aimed at addressing the distribution shift from the class imbalance perspective. However, these approaches oversimplify the issue of distribution shift, leading to suboptimal performance across these three benchmarks.
iCaRL struggles with larger datasets such as Split CIFAR-100/Tiny-ImageNet, suggesting that the NCM classifier relies on accurate and well-separate class means, which is difficult to obtain in online CL on large datasets. A low ACC of iCaRL indicates that it may not effectively acquire available knowledge from each new task, resulting in negligible forgetting during subsequent training and consequently a low FM. As for ER-ACE and SS-IL, although they demonstrate relatively sound decent ACC performance, they excessively suppress the accuracy of new classes during training, which results in a lower FM value. In contrast, the proposed Dual-CBA module balances the ACC and FM and significantly improves the performance of the corresponding baselines.

\begin{table*}[]
\caption{Results of our method under the semi-supervised continual learning setting on three datasets with different buffer sizes. Under this setting, only a small portion (Label Ratio) of the data for each task has labels.}
\label{tab:ssl-results}
\centering
\resizebox{0.95\textwidth}{!}{
\begin{tabular}{cccccccccccccc}
\hline
                        &                                         & \multicolumn{4}{c}{Split CIFAR-10}                                                                                                                                & \multicolumn{4}{c}{Split CIFAR-100}                                                                                                                             & \multicolumn{4}{c}{Split Tiny-ImageNet}                                                                                                                         \\ \cline{3-14} 
                        &                                         & \multicolumn{2}{c}{M=0.2k}                                                      & \multicolumn{2}{c}{M=0.5k}                                                      & \multicolumn{2}{c}{M=2k}                                                     & \multicolumn{2}{c}{M=5k}                                                     & \multicolumn{2}{c}{M=2k}                                                     & \multicolumn{2}{c}{M=5k}                                                     \\ \cline{3-14} 
\multirow{-3}{*}{\makecell[c]{Label\\Ratio}} & \multirow{-3}{*}{Method}                & ACC $\uparrow$                         & FM $\downarrow$                        & ACC $\uparrow$                         & FM $\downarrow$                        & ACC $\uparrow$                         & FM $\downarrow$                        & ACC $\uparrow$                         & FM $\downarrow$                        & ACC $\uparrow$                         & FM $\downarrow$                        & ACC $\uparrow$                         & FM $\downarrow$                        \\ \hline
                        & ER                                  & 38.62                              & 45.84                              & 43.34                              & 39.00                              & 16.38                              & 33.62                             & 19.07                              & 33.52                             & 10.83                              & 27.42                             & 10.68                              & 29.23                             \\
                        & ER-CBA                              & 37.00                              & 45.64                              & 41.95                              & 39.93                              & 15.98                              & 33.86                             & 16.25                              & 34.92                             & 10.07                              & 29.43                             & 10.78                              & 28.70                             \\
                        & \cellcolor{gray!40}ER-Dual-CBA (ours)  & \cellcolor{gray!40}\textbf{41.25} & \cellcolor{gray!40}\textbf{22.62} & \cellcolor{gray!40}\textbf{45.29} & \cellcolor{gray!40}\textbf{17.28} & \cellcolor{gray!40}\textbf{24.67} & \cellcolor{gray!40}\textbf{7.03} & \cellcolor{gray!40}\textbf{26.30} & \cellcolor{gray!40}\textbf{6.06} & \cellcolor{gray!40}\textbf{15.81} & \cellcolor{gray!40}\textbf{6.60} & \cellcolor{gray!40}\textbf{18.79} & \cellcolor{gray!40}\textbf{5.83} \\ \cline{2-14}
                        & DER++                               & 41.02                              & 36.93                              & 46.62                              & 34.43                              & 17.91                              & 31.23                             & 18.73                              & 31.37                             & 10.75                              & 30.40                             & 11.93                              & 28.29                             \\
                        & DER-CBA                             & 41.51                              & 37.30                              & 48.47                              & 26.66                              & 20.84                              & 19.08                             & 22.94                              & 19.12                             & 12.14                              & 21.10                             & 14.97                              & 17.94                             \\
\multirow{-6}{*}{0.2}   & \cellcolor{gray!40}DER-Dual-CBA (ours) & \cellcolor{gray!40}\textbf{42.36} & \cellcolor{gray!40}\textbf{21.03} & \cellcolor{gray!40}\textbf{48.90} & \cellcolor{gray!40}\textbf{9.54} & \cellcolor{gray!40}\textbf{23.32} & \cellcolor{gray!40}\textbf{9.54} & \cellcolor{gray!40}\textbf{27.30} & \cellcolor{gray!40}\textbf{6.58} & \cellcolor{gray!40}\textbf{16.48} & \cellcolor{gray!40}\textbf{7.91} & \cellcolor{gray!40}\textbf{18.53} & \cellcolor{gray!40}\textbf{7.09} \\ \hline
                        & ER                                  & 35.75                              & 44.93                              & 39.04                              & 43.58                              & 15.80                              & 31.64                             & 18.17                              & 29.51                             & 11.27                              & 25.84                             & 11.86                              & 25.22                             \\
                        & ER-CBA                              & 36.21                              & 45.26                              & 41.56                              & 40.09                              & 16.83                              & 30.00                             & 17.41                              & 29.38                             & 10.09                              & 25.62                             & 10.56                              & 26.06                             \\
                        & \cellcolor{gray!40}ER-Dual-CBA (ours)  & \cellcolor{gray!40}\textbf{39.86} & \cellcolor{gray!40}\textbf{22.95} & \cellcolor{gray!40}\textbf{46.87} & \cellcolor{gray!40}\textbf{19.84} & \cellcolor{gray!40}\textbf{22.39} & \cellcolor{gray!40}\textbf{7.30} & \cellcolor{gray!40}\textbf{23.07} & \cellcolor{gray!40}\textbf{7.05} & \cellcolor{gray!40}\textbf{15.15} & \cellcolor{gray!40}\textbf{8.10} & \cellcolor{gray!40}\textbf{17.90} & \cellcolor{gray!40}\textbf{5.82} \\ \cline{2-14}
                        & DER++                               & 40.20                              & 35.56                              & 44.08                              & 33.18                              & 16.79                              & 29.77                             & 20.67                              & 26.46                             & 11.81                              & 26.50                             & 12.33                              & 25.35                             \\
                        & DER-CBA                             & 42.09                              & 31.78                              & 47.20                              & 27.45                              & 21.17                              & 15.84                             & 22.96                              & 14.43                             & 13.73                              & 15.75                             & 14.57                              & 16.48                             \\
\multirow{-6}{*}{0.1}   & \cellcolor{gray!40}DER-Dual-CBA (ours) & \cellcolor{gray!40}\textbf{43.11} & \cellcolor{gray!40}\textbf{18.97} & \cellcolor{gray!40}\textbf{48.78} & \cellcolor{gray!40}\textbf{15.12} & \cellcolor{gray!40}\textbf{22.11} & \cellcolor{gray!40}\textbf{8.74} & \cellcolor{gray!40}\textbf{23.99} & \cellcolor{gray!40}\textbf{5.90} & \cellcolor{gray!40}\textbf{16.00} & \cellcolor{gray!40}\textbf{7.98} & \cellcolor{gray!40}\textbf{18.21} & \cellcolor{gray!40}\textbf{5.88} \\ \hline
\end{tabular}}
\end{table*}

\subsection{Comparison under Semi-Supervised Scenario}
\label{subsec:ssl-CL}
In semi-supervised continual learning, only a small portion of samples from each task is labeled with the proportion of these labeled data referred to as label ratio. We apply the FixMatch~\cite{fixmatch} to two rehearsal-based CL methods, namely ER and DER++. Similar to the fully supervised setting, our method Dual-CBA can also be easily applied to these baselines, and the comparison results across different datasets and various label ratios are shown in \myref{Table~\ref{tab:ssl-results}}. It can be observed that CBA only marginally improves the baselines in some cases, with overall performance showing modest gains. For example, when the label ratio is 0.2, ER-CBA even decreases the ACC and increases the FM of baseline ER across all three datasets. In contrast, our Dual-CBA consistently improves the performance of both baselines and surpasses CBA by a large margin, especially in terms of the FM metric. This is likely because the class-specific CBA requires more labeled data to adapt to changed posterior distribution for each new task, making it prone to overfitting in a semi-supervised CL setting with limited labeled samples. However, the strong transferability of the proposed class-agnostic CBA reduces the need for labeled data and enhances the performance of our Dual-CBA. Additionally, these results indicate that the proposed IBN is effective in a semi-supervised CL setting.

\begin{table}[]
\caption{Results of our method on Split CIFAR-100 under Blurry-10 and Blurry-30 CL settings.}
\label{tab:blurry-results}
\centering
\resizebox{0.95\columnwidth}{!}{
\begin{tabular}{cccccc}
\hline
                           &                                    & \multicolumn{4}{c}{Split CIFAR-100}                                                                                                           \\ \cline{3-6} 
                           &                                    & \multicolumn{2}{c}{M = 2k}                                            & \multicolumn{2}{c}{M = 5k}                                            \\ \cline{3-6} 
\multirow{-3}{*}{Blurry-K} & \multirow{-3}{*}{Method}           & ACC $\uparrow$                    & FM $\downarrow$                   & ACC $\uparrow$                    & FM $\downarrow$                   \\ \hline
                           & ER                                 & 22.72                             & 32.26                             & 23.44                             & 33.20                             \\
                           & ER-CBA                             & 27.82                             & 7.83                              & 28.96                             & 7.06                              \\
                           & \cellcolor{gray!40}ER-Dual-CBA (ours)  & \cellcolor{gray!40}\textbf{29.87} & \cellcolor{gray!40}\textbf{7.08}  & \cellcolor{gray!40}\textbf{32.01} & \cellcolor{gray!40}\textbf{6.38}  \\ \cline{2-6} 
                           & DER++                              & 18.87                             & 42.71                             & 19.47                             & 42.77                             \\
                           & DER-CBA                            & 27.45                             & 13.53                             & 27.56                             & 14.24                             \\
\multirow{-6}{*}{K = 10}   & \cellcolor{gray!40}DER-Dual-CBA (ours) & \cellcolor{gray!40}\textbf{30.66} & \cellcolor{gray!40}\textbf{9.40}  & \cellcolor{gray!40}\textbf{30.65} & \cellcolor{gray!40}\textbf{9.71}  \\ \hline
                           & ER                                 & 26.60                             & 25.37                             & 27.52                             & 25.46                             \\
                           & ER-CBA                             & 28.23                             & 7.08                              & 30.07                             & 6.06                              \\
                           & \cellcolor{gray!40}ER-Dual-CBA (ours)  & \cellcolor{gray!40}\textbf{31.73} & \cellcolor{gray!40}\textbf{5.60}  & \cellcolor{gray!40}\textbf{32.91} & \cellcolor{gray!40}\textbf{4.53}  \\ \cline{2-6} 
                           & DER++                              & 23.41                             & 34.03                             & 23.74                             & 34.14                             \\
                           & DER-CBA                            & 28.04                             & 11.84                             & 29.09                             & 10.95                             \\
\multirow{-6}{*}{K = 30}   & \cellcolor{gray!40}DER-Dual-CBA (ours) & \cellcolor{gray!40}\textbf{31.52} & \cellcolor{gray!40}\textbf{8.30}  & \cellcolor{gray!40}\textbf{31.98} & \cellcolor{gray!40}\textbf{9.20}  \\ \hline
\end{tabular}}
\vspace{-3mm}
\end{table}

\subsection{Comparison on Blurry Scenario}
\label{subsec:task-blurry}
Following~\cite{gradient-selection-blurry, rainbow-blurry}, we adopt the \textit{Blurry-$K$} online CL setting. It simulates a practical situation where task boundaries are unclear, characterized by an overlap across all tasks. Specifically, a fraction ($K\%$) of the training data from one task may appear in other tasks. Here we take $K=10$ and $30$ as an illustration, and the comparison results on Split CIFAR-100 are summarized in \myref{Table~\ref{tab:blurry-results}}.

It can be observed that our Dual-CBA substantially improves the performance of the corresponding baselines and the original CBA module. For example, CBA only improves the ACC of ER by about 1.76\% with 500 buffer samples under the Blurry-10 setting, while the proposed Dual-CBA can further enhance the performance of ER-CBA by about 2.91\%. Additionally, our method significantly reduces the forgetting of the two baselines as shown in \myref{Table~\ref{tab:blurry-results}}. These results demonstrate that the proposed Dual-CBA can be flexibly applied to various rehearsal baseline models and help them adapt to the posterior distribution shift without being perturbed by unclear task boundaries, further verifying the effectiveness of the proposed improvements on CBA.

\subsection{Comparison under the Offline CL }
\label{subsec:offline-CL}
To further validate the generalization ability of the proposed Dual-CBA, we extend our method Dual-CBA to offline continual learning, where each task can be trained over multiple epochs to achieve more stable convergence. As shown in \myref{Table~\ref{tab:offline-results}}, all four baselines achieve better optimum solutions compared to the online context. Obviously, the proposed Dual-CBA significantly improves the performance of corresponding baselines and consistently outperforms the CBA module. These results verify the negative impact of the training bias on CL models and the strength of our method, which can also help the baseline models to adapt to the shifting posterior distribution in the offline setting, indicating the strong generalization ability of the proposed Dual-CBA.

\begin{table}[]
\caption{Results of our method applied on 4 baselines under the offline CL setting on Split CIFAR-100.}
\label{tab:offline-results}
\centering
\begin{tabular}{ccccc}
\hline
                         & \multicolumn{4}{c}{Split CIFAR-100}                               \\ \cline{2-5} 
                         & \multicolumn{2}{c}{M=2k}        & \multicolumn{2}{c}{M=5k}        \\ \cline{2-5} 
\multirow{-3}{*}{Method} & ACC $\uparrow$ & FM $\downarrow$ & ACC $\uparrow$ & FM $\downarrow$ \\ \hline
ER                       & 38.09          & 51.51          & 47.53          & 39.47          \\
ER-CBA                   & 44.71          & 38.19          & 54.95          & \textbf{24.19} \\
\rowcolor{gray!40} 
ER-Dual-CBA (ours)           & \textbf{46.89} & \textbf{35.59} & \textbf{55.25} & 24.23          \\ \hline
DER++                    & 53.01          & 29.23          & 58.34          & 22.50          \\
DER-CBA                  & 52.72          & 22.19          & 58.67          & 15.46          \\
\rowcolor{gray!40} 
DER-Dual-CBA (ours)          & \textbf{54.27} & \textbf{19.73} & \textbf{60.26} & \textbf{14.19} \\ \hline
CLSER                    & 42.77          & 45.72          & 55.83          & 31.10          \\
CLSER-CBA                & 49.09          & \textbf{33.70} & 57.23          & 24.27          \\
\rowcolor{gray!40} 
CLSER-Dual-CBA (ours)        & \textbf{49.20} & 33.87          & \textbf{58.20} & \textbf{21.32} \\ \hline
RAR                      & 48.26          & 34.00          & 54.18          & 26.89          \\
RAR-CBA                  & 50.13          & 25.62          & 56.07          & 19.35          \\
\rowcolor{gray!40} 
RAR-Dual-CBA (ours)          & \textbf{50.43} & \textbf{24.94} & \textbf{56.50} & \textbf{19.12} \\ \hline
\end{tabular}
\end{table}

\subsection{Discussion and Ablation Study.}
\label{subsec:discussion}

\begin{table*}[]
\caption{Accuracy of each task after the whole training process $a_{t, T}~(t=1, \cdots, T)$ and the area under the curve of accuracy $\mathrm{{ACC}_{AUC}}$~\cite{ACC_AUC}. For Split CIFAR-10, task 5 is the new task, and tasks 1-4 are old tasks. For Split CIFAR-100 and Tiny-ImageNet, task 10 is the incoming new task.}
\label{tab:each-task-acc}
\resizebox{\textwidth}{!}{
\begin{tabular}{cccccccccccccc}
\hline 
                                                          & Method                            & \multicolumn{2}{c}{$a_{1,5}$}                       & \multicolumn{2}{c}{$a_{2,5}$}                       & \multicolumn{2}{c}{$a_{3,5}$}                       & \multicolumn{2}{c}{$a_{4,5}$}                       & \multicolumn{2}{c}{$a_{5,5}$}                       & ${\rm{ACC}}$                      & ${\rm{{ACC}_{AUC}}}$              \\ \cline{2-14} 
                                                          & ER                                & \multicolumn{2}{c}{17.02}                           & \multicolumn{2}{c}{15.86}                           & \multicolumn{2}{c}{27.62}                           & \multicolumn{2}{c}{38.30}                           & \multicolumn{2}{c}{82.98}                           & 36.36                             & 33.44                             \\
                                                          & ER-CBA                            & \multicolumn{2}{c}{24.26}                           & \multicolumn{2}{c}{20.78}                           & \multicolumn{2}{c}{25.86}                           & \multicolumn{2}{c}{39.45}                           & \multicolumn{2}{c}{80.29}                           & 38.13                             & 34.96                             \\
\multirow{-4}{*}{\thead{Split CIFAR-10 \\ (M=0.2k)}}      & \cellcolor{gray!40}ER-Dual-CBA (ours) & \multicolumn{2}{c}{\cellcolor{gray!40}44.66}        & \multicolumn{2}{c}{\cellcolor{gray!40}26.19}        & \multicolumn{2}{c}{\cellcolor{gray!40}31.18}        & \multicolumn{2}{c}{\cellcolor{gray!40}40.98}        & \multicolumn{2}{c}{\cellcolor{gray!40}64.00}        & \cellcolor{gray!40}\textbf{41.40} & \cellcolor{gray!40}\textbf{37.10} \\ \hline
                                                          & Method                            & $a_{1,10}$               & $a_{2,10}$               & $a_{3,10}$               & $a_{4,10}$               & $a_{5,10}$               & $a_{6,10}$               & $a_{7,10}$               & $a_{8,10}$               & $a_{9,10}$               & $a_{10,10}$              & ${\rm{ACC}}$                      & ${\rm{{ACC}_{AUC}}}$              \\ \cline{2-14} 
                                                          & ER                                & 19.10                    & 15.13                    & 22.63                    & 13.15                    & 16.70                    & 16.57                    & 21.18                    & 15.55                    & 12.37                    & 64.19                    & 21.66                             & 21.41                             \\
                                                          & ER-CBA                            & 24.03                    & 21.19                    & 29.18                    & 25.01                    & 25.04                    & 31.64                    & 32.05                    & 29.27                    & 15.18                    & 23.03                    & 25.56                             & 22.78                             \\
\multirow{-4}{*}{\thead{Split CIFAR-100 \\ (M=2k)}}       & \cellcolor{gray!40}ER-Dual-CBA (ours) & \cellcolor{gray!40}27.58 & \cellcolor{gray!40}20.73 & \cellcolor{gray!40}33.22 & \cellcolor{gray!40}25.72 & \cellcolor{gray!40}29.37 & \cellcolor{gray!40}36.79 & \cellcolor{gray!40}34.65 & \cellcolor{gray!40}29.59 & \cellcolor{gray!40}31.04 & \cellcolor{gray!40}22.68 & \cellcolor{gray!40}\textbf{29.14} & \cellcolor{gray!40}\textbf{26.34} \\ \hline
                                                          & Method                            & $a_{1,10}$               & $a_{2,10}$               & $a_{3,10}$               & $a_{4,10}$               & $a_{5,10}$               & $a_{6,10}$               & $a_{7,10}$               & $a_{8,10}$               & $a_{9,10}$               & $a_{10,10}$              & ${\rm{ACC}}$                      & ${\rm{{ACC}_{AUC}}}$              \\ \cline{2-14} 
                                                          & ER                                & 11.10                    & 11.10                    & 11.55                    & 13.98                    & 12.30                    & 11.94                    & 10.10                    & 12.39                    & 6.21                     & 49.60                    & 15.03                             & 14.76                             \\
                                                          & ER-CBA                            & 15.80                    & 13.57                    & 17.64                    & 20.04                    & 17.36                    & 16.71                    & 18.24                    & 19.21                    & 12.38                    & 25.96                    & 17.69                             & 15.74                             \\
\multirow{-4}{*}{\thead{Split Tiny-ImageNet \\ (M=2k)}}   & \cellcolor{gray!40}ER-Dual-CBA (ours) & \cellcolor{gray!40}18.35 & \cellcolor{gray!40}14.71 & \cellcolor{gray!40}19.09 & \cellcolor{gray!40}24.65 & \cellcolor{gray!40}19.39 & \cellcolor{gray!40}20.75 & \cellcolor{gray!40}22.67 & \cellcolor{gray!40}22.37 & \cellcolor{gray!40}19.46 & \cellcolor{gray!40}26.67 & \cellcolor{gray!40}\textbf{20.81} & \cellcolor{gray!40}\textbf{18.96} \\ \hline
\end{tabular}}
\end{table*}

\medskip
\noindent\textbf{Dual-CBA helps the model adapt to distribution shifts.}
To ascertain this, we exhibit the accuracy of each task after the final task training, \textit{i.e.}, $a_{t, T}$ (where $t = 1, \cdots, T$) in \myref{Table~\ref{tab:each-task-acc}}. It can be observed that the baseline ER focuses on the new task too much and easily forgets the previously learned knowledge, that is the so-called task-recency bias. However, the proposed Dual-CBA significantly improves the accuracy of previous tasks and prevents the model from paying much more attention to the new task, indicating that our method can help the baseline model adapt to the dramatic distribution shift and absorb the training bias during the continual learning process. Additionally, the Dual-CBA module shows more powerful ability to mitigate task-recency bias than CBA across all datasets, further demonstrating the effectiveness of the proposed enhancement in Dual-CBA.

\begin{figure*}[!t]
\centering
\subfloat[]{
    \includegraphics[height=4.2cm]{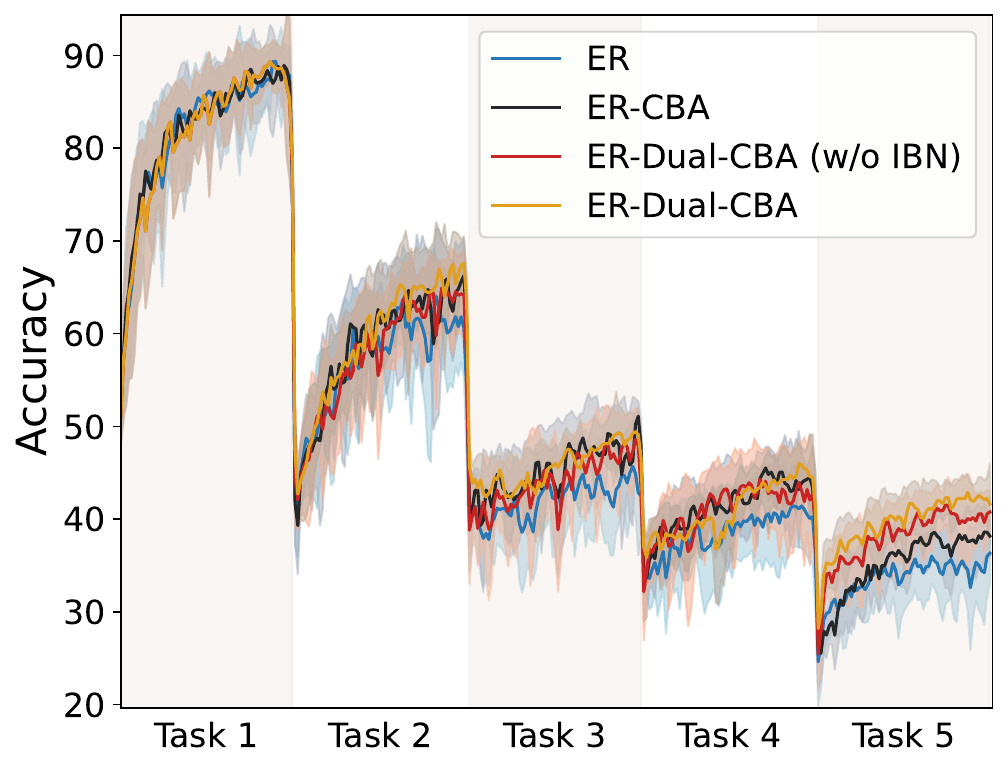}%
    \label{subfig:acc-auc-cifar10}}
\hspace{8pt}
\subfloat[]{
    \includegraphics[height=4.2cm]{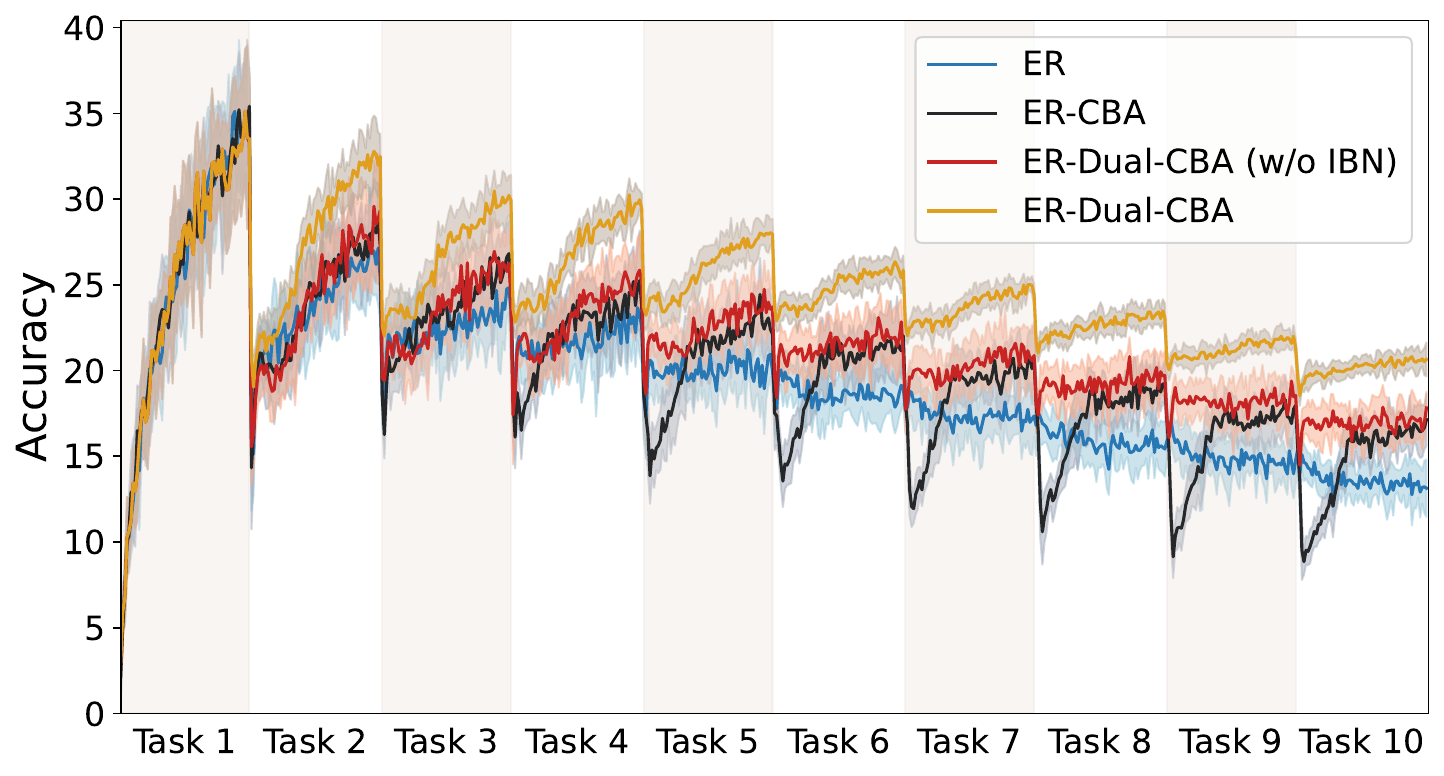}%
    \label{subfig:acc-auc-tinyimg}}
\caption{The average accuracy during the whole continual training process on (a) Split CIFAR-10 with buffer size M=0.2k, and (b) Split Tiny-ImageNet with buffer size M=2k.}
\vspace{-3mm}
\label{fig:acc-auc-cifar10-tinyimg}
\end{figure*}

\medskip
\noindent\textbf{Dual-CBA meets the need for real-time evaluation in Online CL.}
In online continual learning, a challenge is that the model should be evaluated at any time throughout the CL process~\cite{ACC_AUC, real_time_eval}. To explore the effectiveness of our method for any time inference, we display the ACC of the baseline ER, ER-CBA, the proposed ER-Dual-CBA, and ER-Dual-CBA without IBN in \myref{Fig.~\ref{fig:acc-auc-cifar100}} and \myref{Fig.~\ref{fig:acc-auc-cifar10-tinyimg}}. It can be observed that especially on the larger datasets Split CIFAR-100 and Tiny-ImageNet, ER-CBA (\textcolor[RGB]{38,39,41}{black} line) requires a few iterations to adapt the incoming data distribution and then improve the performance of ER (\textcolor[RGB]{40,120,181}{blue} line) quickly. However, ER-Dual-CBA without IBN (\textcolor[RGB]{200,36,35}{red} line) can surpass the baseline ER and ER-CBA at the beginning of each task and keep the higher performance during the entire CL process, which demonstrates that the proposed Dual-CBA can be evaluated in real-time and mitigate the stability-gap problem. Additionally, ER-Dual-CBA equipped with IBN (\textcolor[RGB]{225,159,32}{yellow} line) further illustrates the effectiveness of the proposed IBN, which can significantly assimilate the feature extraction bias during inference. Furthermore, we also calculate $\rm{{ACC}_{AUC}}$ in the last column of \myref{Table~\ref{tab:each-task-acc}}. Our method can significantly improve the $\rm{{ACC}_{AUC}}$ of baseline ER and ER-CBA across different datasets, also indicating that our method can improve the performance of baseline in real-time. Please refer to \myref{Appendix~F} for more experimental results.

\begin{table}[]
\caption{Comparison results on other normalization methods in continual learning on Split CIFAR-100 with different buffer sizes.}
\label{tab:bn-results}
\centering
\begin{tabular}{cccccc}
\hline
                           &                                & \multicolumn{4}{c}{Split CIFAR-100}                                                                       \\ \cline{3-6} 
                           &                                & \multicolumn{2}{c}{M = 0.2k}                        & \multicolumn{2}{c}{M = 0.5k}                        \\ \cline{3-6} 
\multirow{-3}{*}{Method}   & \multirow{-3}{*}{Normalization}& ACC $\uparrow$           & FM $\downarrow$          & ACC $\uparrow$           & FM $\downarrow$          \\ \hline
                           & BN~\cite{BN}                   & 21.66                    & 33.21                    & 23.38                    & 33.19                    \\
                           & CN~\cite{CN}                   & 22.50                    & 34.22                    & 22.44                    & 35.05                    \\
                           & AdaB$^2$N~\cite{AdaB2N}           & 22.40                    & \textbf{31.48}                    & 23.71                    & \textbf{31.94}                    \\
\multirow{-4}{*}{ER}       & \cellcolor{gray!40}IBN (ours) & \cellcolor{gray!40}\textbf{24.67} & \cellcolor{gray!40}33.16 & \cellcolor{gray!40}\textbf{25.34} & \cellcolor{gray!40}33.96 \\ \hline
                           & BN~\cite{BN}                   & 25.29                    & 9.92                     & 28.22                    & 8.77                     \\
                           & CN~\cite{CN}                   & 25.51                    & 9.71                     & 28.40                    & 8.36                     \\
                           & AdaB$^2$N~\cite{AdaB2N}           & 26.37                    & 9.04                     & 28.10                    & 9.16                     \\
\multirow{-4}{*}{ER-Dual-CBA}  & \cellcolor{gray!40}IBN (ours) & \cellcolor{gray!40}\textbf{29.14} & \cellcolor{gray!40}\textbf{8.90}  & \cellcolor{gray!40}\textbf{31.08} & \cellcolor{gray!40}\textbf{7.29}  \\ \hline
                           & BN~\cite{BN}                   & 17.83                    & 43.91                    & 17.36                    & 44.77                    \\
                           & CN~\cite{CN}                   & 18.24                    & 44.23                    & 17.05                    & 45.93                    \\
                           & AdaB$^2$N~\cite{AdaB2N}           & 18.62                    & \textbf{41.35}                    & 18.46                    & \textbf{41.34}                    \\
\multirow{-4}{*}{DER++}    & \cellcolor{gray!40}IBN (ours) & \cellcolor{gray!40}\textbf{18.71} & \cellcolor{gray!40}42.47 & \cellcolor{gray!40}\textbf{18.78} & \cellcolor{gray!40}42.82 \\ \hline
                           & BN~\cite{BN}                   & 25.10                    & 11.02                    & 25.99                    & 10.73                    \\
                           & CN~\cite{CN}                   & 26.04                    & 10.99                     & 25.55                    & 10.60                    \\
                           & AdaB$^2$N~\cite{AdaB2N}           & 26.00                    & 14.30                    & 27.01                    & 13.35                    \\
\multirow{-4}{*}{DER-Dual-CBA} & \cellcolor{gray!40}IBN (ours) & \cellcolor{gray!40}\textbf{27.93} & \cellcolor{gray!40}\textbf{10.24} & \cellcolor{gray!40}\textbf{29.87} & \cellcolor{gray!40}\textbf{8.31}  \\ \hline
\end{tabular}
\end{table}

\medskip
\noindent\textbf{IBN generalizes well to alleviate feature bias.}
To demonstrate the effectiveness of the proposed IBN, we compare it with the traditional batch normalization~\cite{BN}, and two recent normalization methods aiming to adjust the bias introduced by BN, \textit{i.e.}, CN~\cite{CN} and AdaB$^2$N~\cite{AdaB2N}. \myref{Table~\ref{tab:bn-results}} summarizes the comparison results of different normalization methods based on two baselines and the proposed Dual-CBA. It can be observed that: 1) The proposed IBN can be used for both our bi-level optimization framework of Dual-CBA and the corresponding baselines, indicating that IBN is a plug-and-play module, which does not affect the original BN during training and only assimilate the feature extraction bias in the testing stage. 2) CN and AdaB$^2$N can somewhat mitigate the feature extraction bias, while our IBN can significantly improve ACC and reduce FM of corresponding CL baselines, especially when applied to our bi-level optimization framework on Dual-CBA. For example, in the case of memory buffer size M = 0.5k, IBN improves the ACC of baseline DER++ by about 1.42 and further improves that of DER-Dual-CBA by about 3.88. These results demonstrate the effectiveness of the proposed IBN and indicate that the proposed Dual-CBA module and IBN can jointly further improve model performance in continual learning.

\begin{figure*}[!t]
\centering
\subfloat[]{
    \includegraphics[height=4cm]{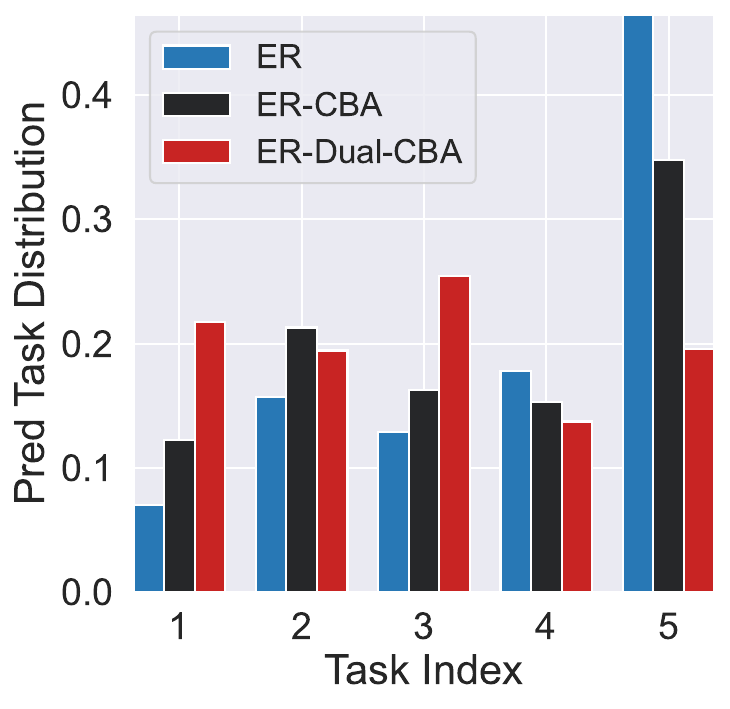}%
    \label{subfig:distribution-cifar10}}
\hspace{4pt}
\subfloat[]{
    \includegraphics[height=4cm]{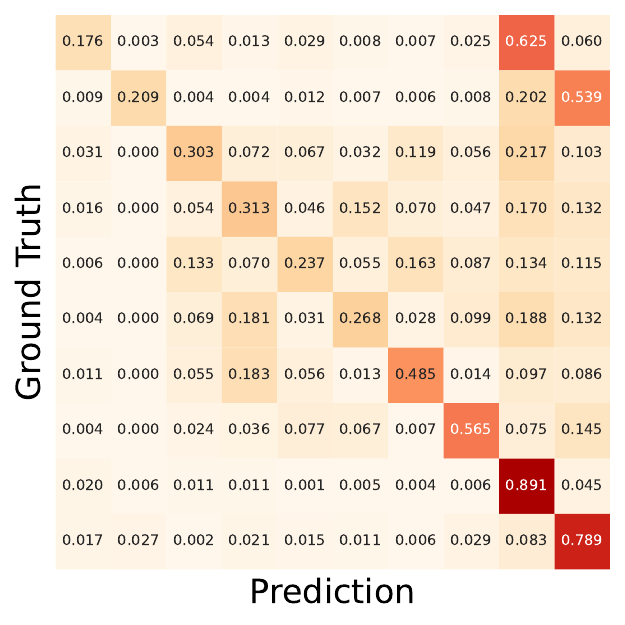}%
    \label{subfig:confusion-er}}
\hspace{4pt}
\subfloat[]{
    \includegraphics[height=4cm]{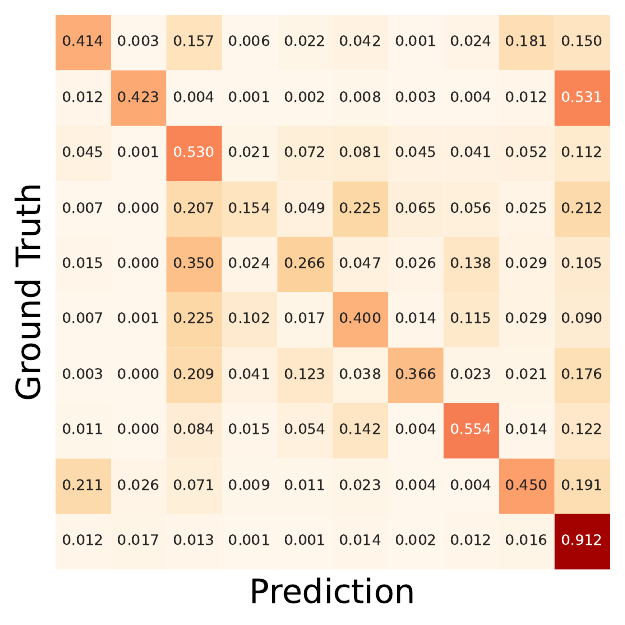}%
    \label{subfig:confusion-ercba}}
\hspace{4pt}
\subfloat[]{
    \includegraphics[height=4cm]{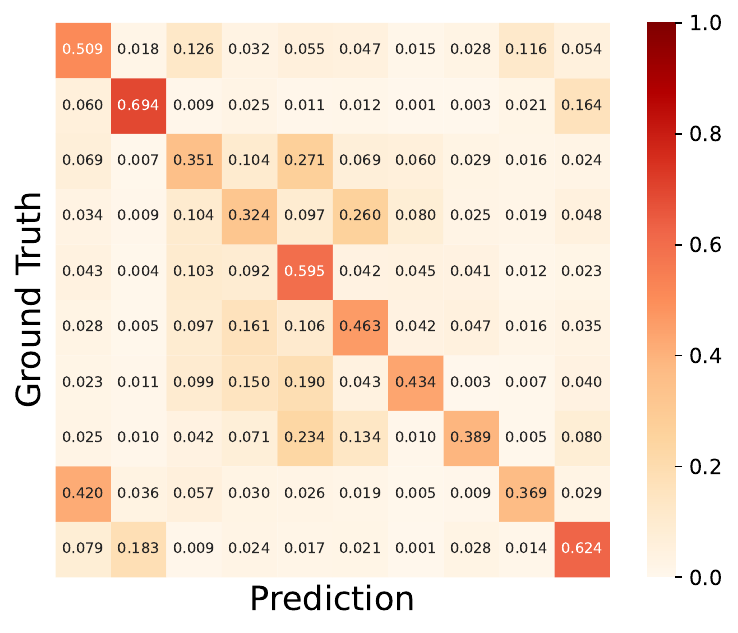}%
    \label{subfig:confusion-erDual-CBA}}
\caption{(a) The prediction distribution of each task on Split CIFAR-10 with buffer size M=0.5k. (b-d) The normalized confusion matrices on Split CIFAR-10 with M=0.5k of ER, ER-CBA, and ER-Dual-CBA, respectively.}
\vspace{-3mm}
\label{fig:task-distrib}
\end{figure*}

\medskip
\noindent\textbf{Comparison of predicted distributions with and without the Dual-CBA module.}
To investigate this discrepancy, we calculate the predicted distribution of each task after training on the final task of Split CIFAR-10 (M=0.5k) in \myref{Fig.~\ref{fig:task-distrib} (a)}. It can be observed that the predicted distribution of the baseline model ER is significantly higher than those of previous tasks, and CBA slightly corrects this task-recency bias. In contrast, our method Dual-CBA can suppress the prediction probability of the new task and automatically adapt to the changing distribution. Furthermore, we also display their confusion matrix on Split CIFAR-10 as shown in \myref{Fig.~\ref{fig:task-distrib} (b-d)}. Obviously, the baseline model ER suffers from a significant task-recency bias, as it achieves an excessively high ACC for new-task classes but is accompanied by extremely severe forgetting. However, CBA only marginally improves old task performances over the baseline model, and the model remains biased toward the new task. Contrastively, the proposed Dual-CBA effectively improves the baseline model to classify samples from all tasks with a global view, leading to the classification network producing a more flatness and correct prediction.

\begin{table}[]
\caption{Intra-dataset transferability of class-agnostic CBA.}
\label{tab:cba-transfer-intra}
\centering
\begin{tabular}{cccc}
\hline
                                      &                                           & \multicolumn{2}{c}{Transfer to other tasks}                           \\ \cline{3-4} 
\multirow{-2}{*}{Dataset}             & \multirow{-2}{*}{Pretrained tasks}        & ACC $\uparrow$                    & FM $\downarrow$                   \\ \hline
                                      & 1st-2nd                                   & 40.41                             & 35.36                             \\
                                      & 1st-3rd                                   & 41.06                             & 30.24                             \\
                                      & 1st-4th                                   & 41.71                             & 29.14                             \\
\multirow{-4}{*}{Split CIFAR-10}      & \cellcolor{gray!40}1st-5th (no transfer)  & \cellcolor{gray!40}\textbf{42.59} & \cellcolor{gray!40}\textbf{24.49} \\ \hline
                                      & 1st-2nd                                   & 28.40                             & 18.75                             \\
                                      & 1st-4th                                   & 28.85                             & 11.88                             \\
                                      & 1st-6th                                   & 29.08                             & \textbf{9.39}                     \\
                                      & 1st-8th                                   & 29.34                             & 9.69                              \\
\multirow{-5}{*}{Split CIFAR-100}     & \cellcolor{gray!40}1st-10th (no transfer) & \cellcolor{gray!40}\textbf{29.81} & \cellcolor{gray!40}9.95           \\ \hline
                                      & 1st-2nd                                   & 18.66                             & 23.20                             \\
                                      & 1st-4th                                   & 19.96                             & 15.86                             \\
                                      & 1st-6th                                   & \textbf{20.93}                    & 12.30                             \\
                                      & 1st-8th                                   & 20.60                             & 11.25                             \\
\multirow{-5}{*}{Split Tiny-ImageNet} & \cellcolor{gray!40}1st-10th (no transfer) & \cellcolor{gray!40}20.55          & \cellcolor{gray!40}\textbf{10.18} \\ \hline
\end{tabular}
\end{table}

\begin{table}[]
\caption{Inter-dataset transferability of class-agnostic CBA.}
\label{tab:cba-transfer-inter}
\centering
\begin{tabular}{ccccc}
\hline
Pretrained dataset                                     & \multicolumn{4}{c}{Transfer to other datasets}                                                                     \\ \hline
\multirow{3}{*}{\thead{Split CIFAR-10\\(M=0.2k)}}      & \multicolumn{2}{c}{\thead{Split CIFAR-100\\(M=2k)}}    & \multicolumn{2}{c}{\thead{Split Tiny-ImageNet\\(M=2k)}}   \\ \cline{2-5} 
                                                       & ACC $\uparrow$       & FM $\downarrow$                 & ACC $\uparrow$        & FM $\downarrow$                   \\
                                                       & 27.15                & 23.25                           & 18.37                 & 24.47                             \\
\rowcolor{gray!40} (no transfer)                       & \textbf{29.14}       & \textbf{8.90}                   & \textbf{20.81}        & \textbf{10.55}                    \\ \hline
\multirow{3}{*}{\thead{Split CIFAR-100\\(M=2k)}}       & \multicolumn{2}{c}{\thead{Split CIFAR-10\\(M=0.2k)}}   & \multicolumn{2}{c}{\thead{Split Tiny-ImageNet\\(M=2k)}}   \\ \cline{2-5} 
                                                       & ACC $\uparrow$       & FM $\downarrow$                 & ACC $\uparrow$        & FM $\downarrow$                   \\
                                                       & \textbf{42.24}       & \textbf{22.26}                  & 20.43                 & \textbf{8.72}                     \\
\rowcolor{gray!40} (no transfer)                       & 41.40                & 28.24                           & \textbf{20.81}        & 10.55                             \\ \hline
\multirow{3}{*}{\thead{Split Tiny-ImageNet\\(M=2k)}}   & \multicolumn{2}{c}{\thead{Split CIFAR-10\\(M=0.2k)}}   & \multicolumn{2}{c}{\thead{Split CIFAR-100\\(M=2k)}}       \\ \cline{2-5} 
                                                       & ACC $\uparrow$       & FM $\downarrow$                 & ACC $\uparrow$        & FM $\downarrow$                   \\
                                                       & \textbf{42.51}       & \textbf{22.82}                  & \textbf{29.77}        & 9.11                              \\
\rowcolor{gray!40} (no transfer)                       & 41.40                & 28.24                           & 29.14                 & \textbf{8.90}                     \\ \hline
\end{tabular}
\end{table}

\medskip
\noindent\textbf{Transferability of the class-agnostic CBA.}
As aforementioned, the proposed class-agnostic CBA has strong transferability and can adapt to different new tasks. To demonstrate this, we conduct experiments to evaluate the intra-dataset and inter-dataset transferability of the class-agnostic CBA module. Concretely, to show the intra-dataset transferability of the proposed CBA, we train the classification network based on ER with the help of the class-agnostic CBA module, which is pretrained in the first few tasks and fixed in the consequent tasks. As shown in \myref{Table~\ref{tab:cba-transfer-intra}}, the pretrained CBA can also assimilate the training bias of the classification network to achieve comparable performance to the on-the-fly training version (where the CBA module is trained simultaneously with the classification network, as shown in the gray cell of \myref{Table~\ref{tab:cba-transfer-intra}}). On the other hand, we test the inter-dataset transferability of the proposed class-agnostic CBA. Specifically, we train the classification network based on ER with the help of a fixed class-agnostic CBA, which has been pre-trained in another dataset. In \myref{Table~\ref{tab:cba-transfer-inter}}, even for various new tasks from different datasets, the class-agnostic CBA module also largely improves the performance of the classification framework. Particularly, after the class-agnostic CBA is trained on a dataset with more tasks such as Split CIFAR-100 and Tiny-ImageNet, it can be well transferred to other datasets and even has better results for Split CIFAR-10, indicating that the class-agnostic CBA can capture the stable relationship between the new and old posterior probabilities across multiple continual learning tasks and effectively generalized to other datasets. In summary, these experimental results demonstrate the strong transferability of the proposed class-agnostic CBA module, indicating the potential to handle complicated relationships between new and old tasks.

\begin{table}[]
\caption{Ablation study on the bi-level optimization framework based on ER-Dual-CBA.}
\label{tab:bilevel}
\begin{tabular}{cccccc}
\hline
\multirow{3}{*}{Optimization} & \multirow{3}{*}{Loss function}             & \multicolumn{4}{c}{Split CIFAR-100}                                 \\ \cline{3-6} 
                              &                                            & \multicolumn{2}{c}{M = 2k}       & \multicolumn{2}{c}{M = 5k}       \\ \cline{3-6} 
                              &                                            & ACC $\uparrow$ & FM $\downarrow$ & ACC $\uparrow$ & FM $\downarrow$ \\ \hline
Single level                  & $\mathcal{L}^{trn}$                        & 24.57          & 34.41           & 25.64          & 34.05           \\
Single level                  & $\mathcal{L}^{trn} + \mathcal{L}^{buf}$    & 24.59          & 34.46           & 25.72          & 34.14           \\
\rowcolor{gray!40} 
Bi-level (ours)               & $\mathcal{L}^{trn}$ \& $\mathcal{L}^{buf}$ & \textbf{29.14} & \textbf{8.90}   & \textbf{31.08} & \textbf{7.29}   \\ \hline        
\end{tabular}
\end{table}

\medskip
\noindent\textbf{Ablation on the bi-level optimization framework.}
In our algorithm, we propose a bi-level learning framework to jointly optimize the parameters of the main classification network and the Dual-CBA module. Here we compare the results of the bi-level and single-level optimization to demonstrate the effectiveness of the proposed bi-level learning framework. In \myref{Table~\ref{tab:bilevel}}, we display two single-level optimization formulations: 1) only use the rehearsal training loss (\textit{i.e.}, the inner objective of \myref{Eq.~(\ref{eq:inner-loss})}) to optimize $\theta$, $\phi$ together; and 2) iteratively optimize $\theta$ by the inner loss and optimize $\phi$ by the outer loss. It can be observed that these two single-level learning cannot enhance the performance of the baseline model. The learned Dual-CBA module optimized in single-level frameworks cannot help the model alleviate the distribution shift, which demonstrates that the proposed bi-level optimization is essential for our model training. Under the bi-level learning framework, Dual-CBA can effectively assimilate the training bias and help the model learn an implicit stable posterior distribution with a consolidation of the seen knowledge.

\section{Conclusion}
\label{sec:conclusion}
In this paper, we address the challenges of online continual learning, where the classification network commonly exhibits dramatically changing posterior distributions, which yields performance bias toward the arriving new tasks. To tackle this, we propose a novel bi-level optimization framework that enables the classification network to adapt to these catastrophic distribution shifts. Specifically, we introduce a Dual-CBA module that includes a class-specific CBA module, a class-agnostic module, and an Incremental BN module. This combined module is capable of adapting to catastrophic distribution shifts and mitigating the stability gap issue arising from continual training. This ensures the classification network meets the requirements for real-time inference in online CL. The effectiveness of our method is demonstrated through both theoretical and empirical evidence. Theoretically, we provide an explanation of how our approach addresses task-recency bias and alleviates catastrophic forgetting, and we offer insights into the bi-level optimization framework through a linear formulation. Empirically, our algorithm consistently improves performance across various rehearsal-based baselines, showcasing its ability to assimilate training bias and effectively consolidate learned knowledge in online CL. Currently, our method is limited to being a plug-in for rehearsal-based methods. In the future, we will further investigate the applicability of our approach to rehearsal-free models.

{
	\bibliographystyle{IEEEtran}
	\bibliography{manu_bib}
}

{\appendix
	\subsection{Details of the Bi-level Optimization}
	\label{appendix:bilevel-details}
	In this section, we first review the proposed bi-level learning framework to optimize the parameters of the classifier network and the Dual-Continual Bias Adaptor (Dual-CBA) module. Formally, the proposed bi-level optimization problem can be formulated as
	\begin{equation}
		\begin{split}
			\min_{\phi} \quad &\mathcal{L}^{buf} \left( \mathcal{B}^{buf}; f_{\theta(\phi)} \right) \\
			{\rm s.t.} \quad &\theta(\phi) = 
			\arg\min_{\theta} \mathcal{L}^{trn} \left( \mathcal{B}^{trn}; \mathcal{F}_{\theta, \phi} \right).
		\end{split}
		\label{eq:bi-level}
	\end{equation}
	We use a nested gradient-optimization-based method to update the classifier network parameter $\theta$ and the Dual-CBA parameter $\phi$. Specifically, in the inner loop, the updating formulation of $\theta$ at iteration step $k$ can be expressed as
	\begin{equation}
		\theta^{k+1}(\phi) = \theta^{k} - \alpha \cdot \nabla_{\theta} \mathcal{L}^{trn} \left(\mathcal{B}^{trn}; \mathcal F_{\theta^k, \phi^k} \right),
		\label{eq:inner}
	\end{equation}
	where $\alpha > 0$ is the learning rate of the inner loop.
	
	Then we present details of the updating formulation of the Dual-CBA parameter $\phi$ in the outer loop. The updating of $\phi$ can be represented as
	\begin{equation}
		\phi^{k+1} = \phi^{k} - \beta \cdot \nabla_{\phi} \mathcal{L}^{buf} \left( \mathcal{B}^{buf}; f_{\theta^{k+1}(\phi)} \right),
		\label{eq:outer}
	\end{equation}
	where $\beta > 0$ is the learning rate. In \myref{Eq.~(\ref{eq:outer})}, the derivation term can be represented as
	\begin{equation}
		\begin{split}
			&\quad \nabla_{\phi} \mathcal{L}^{buf} \left( \mathcal{B}^{buf}; f_{\theta^{k+1}(\phi)} \right) \\
			&= 
			\frac{\partial \mathcal{L}^{buf} \left( \mathcal{B}^{buf}; f_{\theta^{k+1}(\phi)} \right)}{\partial \theta^{k+1}(\phi)} 
			\frac{\partial \theta^{k+1}(\phi)}{\partial \phi} \\
			&= 
			\frac{\partial \mathcal{L}^{buf} \left( \mathcal{B}^{buf}; f_{\theta^{k+1}(\phi)} \right)}{\partial \theta^{k+1}(\phi)} 
			\frac{\partial}{\partial \phi} 
			\left(- \alpha \frac{\partial \mathcal{L}^{trn} \left( \mathcal{B}^{trn}; \mathcal F_{\theta^k, \phi^k} \right)}{\partial \theta} \right) \\
			&= - \alpha 
			\frac{\partial \mathcal{L}^{buf} \left( \mathcal{B}^{buf}; f_{\theta^{k+1}(\phi)} \right)}{\partial \theta^{k+1}(\phi)} 
			\frac{\partial^2 \mathcal{L}^{trn} \left( \mathcal{B}^{trn}; \mathcal F_{\theta^k, \phi^k} \right)}{\partial \phi \partial \theta}.
		\end{split}
		\label{eq:2nd-order}
	\end{equation}
	Obviously, the update of $\phi$ introduces a second-order derivation that can be easily implemented by PyTorch \myref{[66]}. To alleviate the calculation burden of this second-order derivation in \myref{Eq.~(\ref{eq:2nd-order})}, we assume that this derivation only depends on the parameter of the last linear classification layer rather than the whole classifier network like ResNet-18. The linear classification layer only introduces a small number of parameters which can significantly speed up the computation and make our bi-level optimization efficient and suitable for online CL.

	\subsection{Proof and Extension of Theoretical Analysis}
	\label{appendix:theorem-proof}
	
	\subsubsection{Proof of \myref{Theorem 1}}
	\label{appendix-sub:proof-alignment}
	\myref{Theorem 1} illustrates the proposed bi-level optimization of Dual-CBA implicitly aligns the gradient w.r.t $\theta$ on the training set with Dual-CBA and that on the memory buffer without Dual-CBA. Additionally, Dual-CBA can prevent bias from disturbing the gradient of the training set, resulting in a more accurate gradient alignment to alleviate forgetting. In this section, we provide the proof of this theorem in the following.
	
	\noindent
	\textbf{Proof:}
	The bi-level learning framework is represented as \myref{Eq.~(\ref{eq:bi-level})}.
	If the inner optimization problem is approximated by one-step gradient descent by \myref{Eq.~(\ref{eq:inner})}, the $(k+1)$th iteration becomes
	\begin{equation}
		\begin{split}
			\min_{\phi} \quad &\mathcal{L}^{buf} \left( \mathcal{B}^{buf}; f_{\theta^{k+1}(\phi)} \right) \\
			{\rm s.t.} \quad &\theta^{k+1}(\phi)
			= \theta^{k}(\phi) 
			- \alpha \frac{\partial \mathcal{L}^{trn} \left(\mathcal{B}^{trn}; \mathcal{F}_{\theta^{k}, \phi} \right)}{\partial \theta^{k}}.
		\end{split}
		\label{eq:th1_reformulate}
	\end{equation}
	For simplicity in this proof, we omit the dataset of each loss function, \textit{i.e.}, the training loss of the inner and outer loop can be reformulated as $\mathcal{L}^{buf} \left(f_{\theta(\phi)} \right)$ and $\mathcal{L}^{trn} \left(\mathcal{F}_{\theta, \phi} \right)$, respectively. 
	
	Assume that the outer-loop training loss function $\mathcal{L}^{buf} \left( f_{\theta(\phi)} \right)$ is $\eta$ gradient Lipschitz w.r.t $\theta$, then we have
	\begin{equation}
		\begin{split}
			&\mathcal{L}^{buf} \left( f_{\theta^{k+1}(\phi)} \right) \\
			\leq &\mathcal{L}^{buf} \left( f_{\theta^{k}(\phi)} \right)
			+ \left\langle
			\frac{\partial \mathcal{L}^{buf} \left( f_{\theta^{k}(\phi)} \right)}{\partial \theta^{k}},
			\theta^{k+1}(\phi) - \theta^{k}(\phi)
			\right\rangle \\
			&\hspace{19.5mm} + \frac{\eta}{2} \left\| \theta^{k+1}(\phi) - \theta^{k}(\phi) \right\|^2
			\\
			= &\mathcal{L}^{buf} \left( f_{\theta^{k}(\phi)} \right) 
			+ \left\langle
			\frac{\partial \mathcal{L}^{buf} \left( f_{\theta^{k}(\phi)} \right)}{\partial \theta^{k}},
			- \alpha \frac{\partial \mathcal{L}^{trn} \left( \mathcal{F}_{\theta^{k}, \phi} \right)}{\partial \theta^{k}}
			\right\rangle \\
			&\hspace{19.5mm} + \frac{\eta}{2}
			\left\|
			\alpha \frac{\partial \mathcal{L}^{trn} \left( \mathcal{F}_{\theta^{k}, \phi} \right)}{\partial \theta^{k}}
			\right\|^2.
		\end{split}
		\label{eq:lipschitz}
	\end{equation}
	In \myref{Eq.~(\ref{eq:lipschitz})}, the inequality on the second line holds due to the $\eta$ gradient Lipschitz property, and the equality on the third line holds due to the constraints in \myref{Eq.~(\ref{eq:th1_reformulate})}.
	
	The update of $\phi$ by SGD as
	\begin{equation}
		\phi^{k+1} = \phi^{k} - \beta \frac{\partial \mathcal{L}^{buf} \left( f_{\theta(\phi)} \right)}{\partial \phi^{k}},
		\nonumber
	\end{equation}
	which aims to decrease the objective $\mathcal{L}^{buf} \left( f_{\theta(\phi)} \right)$ such that $\mathcal{L}^{buf} \left( f_{\theta^{k+1}(\phi)} \right) \leq \mathcal{L}^{buf} \left( f_{\theta^{k}(\phi)} \right)$. According to \myref{Eq.~(\ref{eq:lipschitz})}, one obtains that
	\begin{equation}
		\begin{split}
			& \left\langle
			\frac{\partial \mathcal{L}^{buf} \left(f_{\theta^k(\phi^{k+1})} \right)}{\partial \theta^k}
			\right., \left.
			\frac{\partial \mathcal{L}^{trn} \left(\mathcal{F}_{\theta^k, \phi^{k+1}} \right)}{\partial \theta^k}
			\right\rangle \\
			&\geq
			\frac{\alpha \eta}{2}
			\left\|
			\frac{\partial \mathcal{L}^{trn} \left(\mathcal{F}_{\theta^k, \phi^{k+1}} \right)}{\partial \theta^k}
			\right\|_2^2,
		\end{split}
	\end{equation}
	where $\alpha > 0$ is the inner-loop learning rate and $\eta > 0$ is the Lipschitz constant.

	\medskip
	\subsubsection{Proof of \myref{Theorem 2}}
	\label{appendix-sub:proof-linear}
	In \myref{Theorem 2}, we reformulate the proposed bi-level optimization in a linear version and provide the closed-form solution of $\theta$ and $\phi$. The derivations are summarized as follows.
	
	\noindent
	\textbf{Proof:} Thanks to the excellent properties of convex optimization, we can differentiate the inner loss with respect to $\theta_t$ and let the derivative term be 0:
	\begin{equation}
		\begin{aligned}
			&\quad \frac{\partial}{\partial \theta_t} \left(
			\frac{1}{n^{trn}_t} \lVert (X^{trn}_t)^T \theta_t \phi^*_t - Y^{trn}_t \rVert^2_F
			\right) \\
			&= \frac{1}{n^{trn}_t}
			\frac{\partial}{\partial \theta_t} \mathrm{tr} 
			\Big(
			\left[ (X^{trn}_t)^T \theta_t \phi^*_t - Y^{trn}_t \right]^T  \\
			&\hspace{22.5mm}
			\left[ (X^{trn}_t)^T \theta_t \phi^*_t - Y^{trn}_t \right]
			\Big) \\
			&= \frac{2}{n^{trn}_t} X^{trn}_t \left( (X^{trn}_t)^T \theta_t \phi^*_t - Y^{trn}_t \right) (\phi^*_t)^T = 0 \\
		\end{aligned}
		\label{eq-theta-gradient}
	\end{equation}
	Assuming that $\phi^* \in \mathbb{R}^{c_t \times c_t}$ is a non-singular square matrix, we have 
	\begin{equation}
		\begin{aligned}
			\theta^*_t &= 
			\left( X^{trn}_t (X^{trn}_t)^T \right)^{-1} X^{trn}_t
			Y^{trn}_t (\phi^*_t)^T \left( (\phi^*_t) (\phi^*_t)^T \right)^{-1} \\
			&=  (X^{trn}_t)^\dagger ~
			Y^{trn}_t ~
			(\phi^*_t)^T
			\left( (\phi^*_t)^T \right)^{-1}
			\left(  \phi^*_t    \right)^{-1} \\
			&=  (X^{trn}_t)^\dagger ~ Y^{trn}_t ~ (\phi^*_t)^{-1},
		\end{aligned}
		\label{eq-theta-solution}
	\end{equation}
	where $(X)^\dagger = (XX^T)^{-1} X$.
	
	With \myref{Eq.~(\ref{eq-theta-solution})}, we substitute it to the outer loss in \myref{Eq.~(16)}.
	In this way, the bi-level optimization can be transformed into a convex optimization as follows:
	\begin{equation}
		\begin{aligned}
			\phi^*_t &= \arg\min_\phi \frac{1}{n^{buf}_t} \lVert (X^{buf}_t)^T \theta^*_t(\phi) - Y^{buf}_t \rVert^2_F \\
			&= \arg\min_\phi \frac{1}{n^{buf}_t} 
			\lVert  (X^{buf}_t)^T (X^{trn}_t)^\dagger Y^{trn}_t \phi_t^{-1} - Y^{buf}_t \rVert^2_F \\
			&= \arg\min_\phi \frac{1}{n^{buf}_t} 
			\lVert A Y^{trn}_t \phi_t^{-1} - Y^{buf}_t \rVert^2_F,
		\end{aligned}
		\label{eq-outer-linear}
	\end{equation}
	where we denote the matrix $A = (X^{buf}_t)^T (X^{trn}_t)^\dagger = (X^{buf}_t)^T \left( X^{trn}_t (X^{trn}_t)^T \right)^{-1} X^{trn}_t \in \mathbb{R}^{m_t \times n_t}$ for notation simplicity. $m_t$ and $n_t$ are the sample number of $X^{buf}$ and $X^{trn}$, respectively.
	
	To solve this optimization problem, we differentiate \myref{Eq.~(\ref{eq-outer-linear})} with respect to $\phi_t$,
	\begin{small}
		\begin{equation}
			\begin{aligned}
				&\quad \frac{\partial}{\partial \phi_t} \left(
				\frac{1}{n^{buf}_t} \lVert A Y^{trn}_t \phi_t^{-1}
				- Y^{buf}_t \rVert^2_F
				\right) \\
				&= \frac{1}{n^{buf}_t}
				\frac{\partial}{\partial \phi_t} \mathrm{tr}
				\Big(
				\left[ A Y^{trn}_t \phi_t^{-1} - Y^{buf}_t \right]^T
				\left[ A Y^{trn}_t \phi_t^{-1} - Y^{buf}_t \right]
				\Big) \\
				&= \frac{1}{n^{buf}_t}
				\frac{\partial}{\partial \phi_t}
				\Big[
				\mathrm{tr} \Big( \left[ (\phi_t^{-1})^T (Y^{trn}_t)^T A^T A Y^{trn}_t \phi_t^{-1})^T \right] \Big) \\
				&\hspace{15mm} - \mathrm{tr} \Big( \left[ (\phi_t^{-1})^T (Y^{trn}_t)^T A^T Y^{buf}_t \right] \Big) \\
				&\hspace{15mm} - \mathrm{tr} \Big( \left[ Y^{buf}_t A Y^{trn}_t \phi_t^{-1} \right] \Big)
				\Big]\\
				&= \frac{2}{n^{buf}_t}
				\phi_t^{-T}
				\Big[
				(Y^{buf}_t)^T A Y^{trn}_t - \phi_t^{-T} (Y^{trn}_t)^T A^T A Y^{trn}_t
				\Big]^T
				\phi_t^{-T},
			\end{aligned}
			\label{eq-phi-gradient}
		\end{equation}
	\end{small}
	where $\phi_t^{-T}$ represents the transposed inverse matrix of $\phi$.
	Let \myref{Eq.~(\ref{eq-phi-gradient})} equal zero, we have the solution of $\phi_t$, \textit{i.e.},
	\begin{equation}
		\phi^*_t = 
		\left[ (Y^{trn}_t)^T A^T Y^{buf}_t \right]^{-1}
		\left[ (Y^{trn}_t)^T A^T A Y^{trn}_t \right].
	\end{equation}
	
	Additionally, we substitute $\phi^*$ back to \myref{Eq.~(\ref{eq-theta-solution})}, and we have:
	\begin{equation}
		\begin{aligned}
			\theta^*_t &= 
			(X^{tr}_t)^\dagger
			Y^{tr}_t
			\left[ (Y^{trn}_t)^T A^T A Y^{trn}_t \right]^{-1}
			\left[ (Y^{trn}_t)^T A^T Y^{buf}_t \right].
		\end{aligned}
	\end{equation}

	\begin{table}[]
		\caption{Ablation study on the architecture of the Dual-CBA module.}
		\label{tab:cba-arch}
		\centering
		\resizebox{\columnwidth}{!}{
			\begin{tabular}{ccccccc}
				\hline
				&                                  &                                  & \multicolumn{4}{c}{Split CIFAR-100}                                 \\ \cline{4-7} 
				&                                  &                                  & \multicolumn{2}{c}{M=2k}         & \multicolumn{2}{c}{M=5k}         \\ \cline{4-7} 
				\multirow{-3}{*}{Method} & \multirow{-3}{*}{\makecell[c]{Class-\\specific}} & \multirow{-3}{*}{\makecell[c]{Class-\\agnostic}} & ACC $\uparrow$ & FM $\downarrow$ & ACC $\uparrow$ & FM $\downarrow$ \\
				ER-Dual-CBA-1                & /                                & \checkmark                       & 29.76          & 9.97            & 31.76          & 8.66            \\
				ER-Dual-CBA-2                & single                           & /                                & 28.79          & 10.7            & 30.43          & 9.85            \\
				ER-Dual-CBA-3                & individual                       & /                                & 29.31          & \textbf{9.76}   & 31.63          & 8.57            \\
				\rowcolor{gray!40} 
				ER-Dual-CBA (ours)           & individual                       & \checkmark                       & \textbf{29.81} & 9.95            & \textbf{31.86} & \textbf{8.25}   \\ \hline
		\end{tabular}}
	\end{table}

	\subsection{Implementation Details of Dual-CBA}
	\label{appendix:implementaion}
	\wqza{This section provides implementation details of the proposed Dual-CBA module.}
	\wqza{We use two-layer MLP as the class-specific CBA and the class-agnostic CBA. As we aforementioned in \myref{Sec. III-C}, 
		the class-specific CBA in continual learning may aggregate the stability gap, we reinitialize the class-specific CBA modules to avoid incorrect adjustments when a new task arrives, and then we train it together with the class-agnostic CBA.}
	
	\wqza{In practice, to further simplify the optimization of Dual-CBA, we introduce a skip-layer connection between the outputs of classification network $f_\theta$ and the outputs of Dual-CBA $g_{\phi}$, \textit{i.e.}, the adapted posterior distribution $\tilde{y} = \frac{1}{Z} (g_{\phi}(\hat{y}) + \hat{y})$ where $Z$ is a normalization constant. As verified in previous works~\myref{[74]},
		the skip-layer connection can aid the model convergence and facilitate the gradient backward propagation.}
	
	\wqza{\textit{Optional:} In our experiments, we empirically found that performance can be further improved by using two individual class-specific CBA modules: one for the posterior probabilities of the classes from the new task and another for the classes from all old tasks. This approach is more effective in avoiding the interference between the new and old tasks than using a single class-specific CBA for the posterior probabilities of all new and old classes. To demonstrate this, we conduct ablation studies in \myref{Table~\ref{tab:cba-arch}}.}
	
	In \myref{Table~\ref{tab:cba-arch}}, we compare the effectiveness of each component of the Dual-CBA module, including the two individual class-specific CBA, a single class-specific CBA, and the proposed class-agnostic CBA. ER-Dual-CBA-1 shows the effectiveness of the proposed class-agnostic CBA, which can significantly assimilate the training bias in CL and achieve comparable results to ER-Dual-CBA. Additionally, we set two individual class-specific CBA modules to adapt to the posterior probability of the old and new classes, respectively. Comparing ER-Dual-CBA-2 and 3 in \myref{Table~\ref{tab:cba-arch}}, the performance of two individual class-specific CBA modules is significantly higher than that of only using one class-specific CBA for all classes, indicating that the individual class-specific CBA can adjust the posterior probability more accurate and assimilate the negative influence between new and old tasks. Finally, combining individual class-specific CBA and class-agnostic CBA, our ER-Dual-CBA achieves the best performance and the results demonstrate the effectiveness of each component of our method.

	\subsection{Details of Experiments}
	\label{appendix:details}

	\medskip
	\subsubsection{Comparison Methods}
	\label{appendix-sub:details-comparison}
	We herein detail the four baselines: ER, DER++~\myref{[7]}, RAR~\myref{[8]}, and CLSER~\myref{[25]}; and five state-of-the-art comparison methods presented in the main text: iCaRL~\myref{[6]}, LUCIR~\myref{[11]}, BiC~\myref{[12]}, ER-ACE~\myref{[13]}, and SS-IL~\myref{[10]}.
	\begin{itemize}
		\item \textbf{ER} is the most commonly used baseline in the continual learning problem, which has been introduced in \myref{Sec.~3.1}. 
		\item \textbf{DER++}~\myref{[7]} additionally saves the output logits in the memory buffer and utilizes an additional logit distillation to further prevent forgetting the previously learned knowledge, which is another simple yet strong and effective baseline. 
		\item \textbf{RAR}~\myref{[8]} applies random augmentation to prevent overfitting on the memory buffer samples, which is also a plug-and-play strategy that can be used for many rehearsal-based CL methods. In our experiments, we apply RAR on DER++ which is also noted as RAR for simplicity.
		\item \textbf{CLSER}~\myref{[25]} involves a stable network and a plastic network to consolidate the previous knowledge and learn recent experiences, respectively. 
		\item \textbf{iCaRL}~\myref{[6]} replaces the linear classifier as the nearest class mean (NCM) classifier during inference and chooses samples that are close to class means as the memory buffer. This buffer management strategy facilitates the calculation of more accurate class means. So we still use this strategy for a better prediction performance, although the greedy algorithm does not fully meet the online setting.
		\item \textbf{LUCIR}~\myref{[11]} proposes a weight normalization strategy to rebalance the new and old samples. It is also coupled with a cosine constraint and a margin ranking loss for negative sample learning.
		\item \textbf{BiC}~\myref{[12]} splits a small balanced validation set from the whole training data used for the second stage training. Following the origin paper, the ratio of train/validation split on the examples is also set as 9:1 in our experiments. In the second stage, we train the bias correction layer for 250 epochs by the Adam optimizer with a learning rate of 0.001. 
		\item \textbf{ER-ACE}~\myref{[13]} separates the losses on the incoming new task data and the buffered old task data and only considers news classes for the denominator of CE loss on the incoming data. This asymmetric update pushes new classes to adapt to the older ones.
		\item \textbf{SS-IL}~\myref{[10]} proposes a separated softmax blocking the flow of the gradients between the old and new classes, which can avoid the imbalance penalization for the new and old samples. An additional task-specific distillation is used on the rehearsal data.
	\end{itemize}

	\begin{figure}[t]
		\centering
		\subfloat[]{
			\includegraphics[width=0.75\columnwidth]{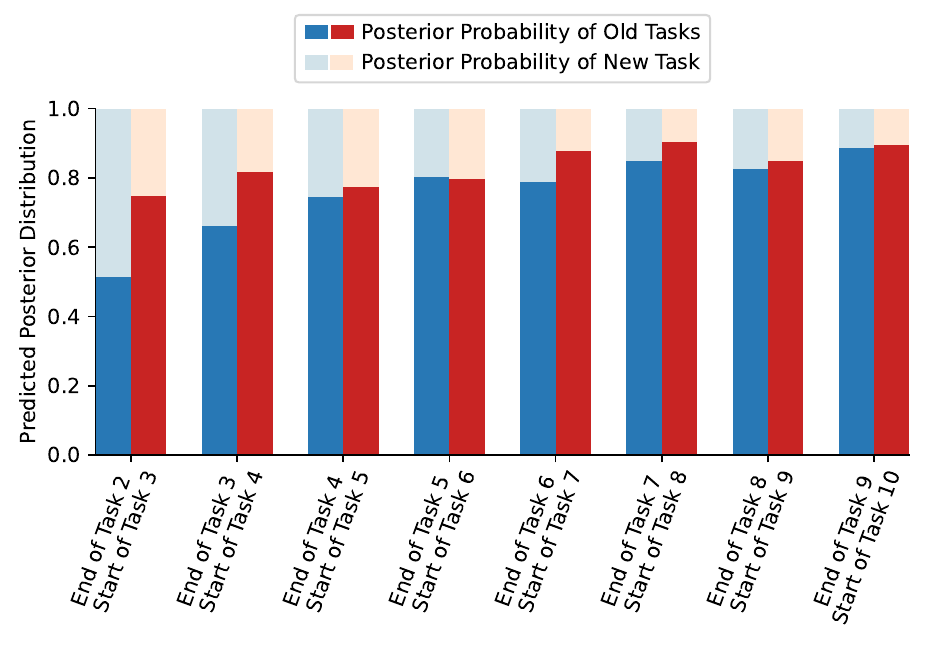}%
			\label{subfig:distrib_change_tsk_old}}
		\hspace{8pt}
		\subfloat[]{
			\includegraphics[width=0.75\columnwidth]{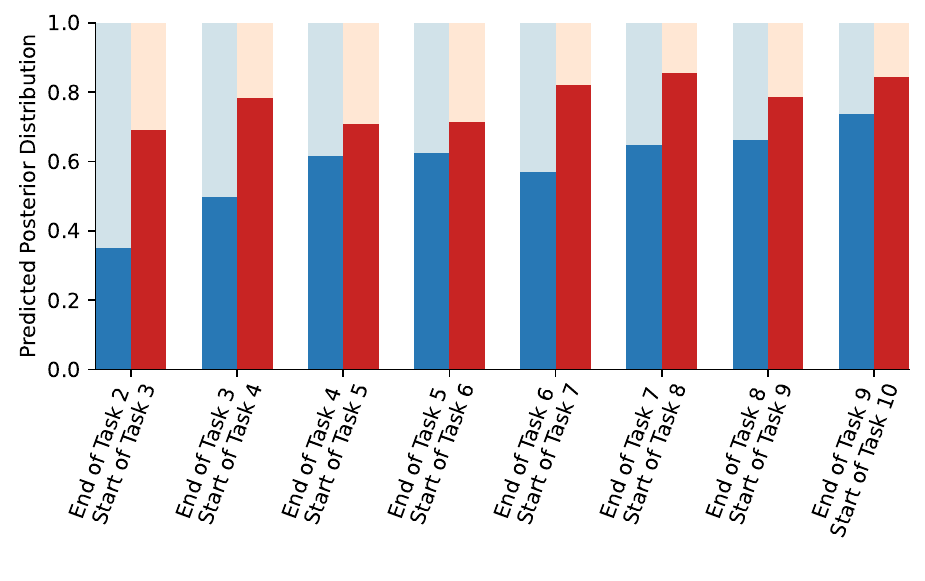}%
			\label{subfig:distrib_change_tsk_new}}
		\caption{\wqza{Illustration of the old-task posterior probability $\hat{y}_{old}$ and the new-task posterior probability $\hat{y}_{new}$ predicted by the classification network at each task transition timestamp during the whole continual learning process: (a) test samples from all old tasks; (b) test samples from the new task. The classification network is trained with the help of our proposed Dual-CBA on Split CIFAR-100.}}
		\vspace{-3mm}
		\label{fig:distrib_change_tsk_dcba}
	\end{figure}

	\medskip
	\subsubsection{Experiment Details}
	\label{appendix-sub:details-experiment}
	In this section, we detail the specific hyper-parameters setting of different methods and datasets in our experiment.
	
	\medskip
	\noindent
	\textbf{Details of the datasets.} For CIFAR-10 and CIFAR-100~\myref{[15]}, both of them consist of 50,000 32 $\times$ 32 RGB images for training and 10,000 test images, whereas CIFAR-10 only contains 10 classes and CIFAR-100 contains 100 classes. We evenly split them into 5 tasks and 10 tasks in order, respectively, which are noted as Split CIFAR-10 and Split CIFAR-100.
	In addition, for the Tiny-ImageNet~\myref{[73]}, there are 100,000 color images of 200 different classes cropped as 64$\times$64. Similarly, we split the Tiny-ImageNet into 10 tasks (\textit{i.e.}, Split Tiny-ImageNet), and each task contains 20 disjoint classes.

	\begin{figure*}[t]
		\begin{center}
			\includegraphics[width=0.9\linewidth]{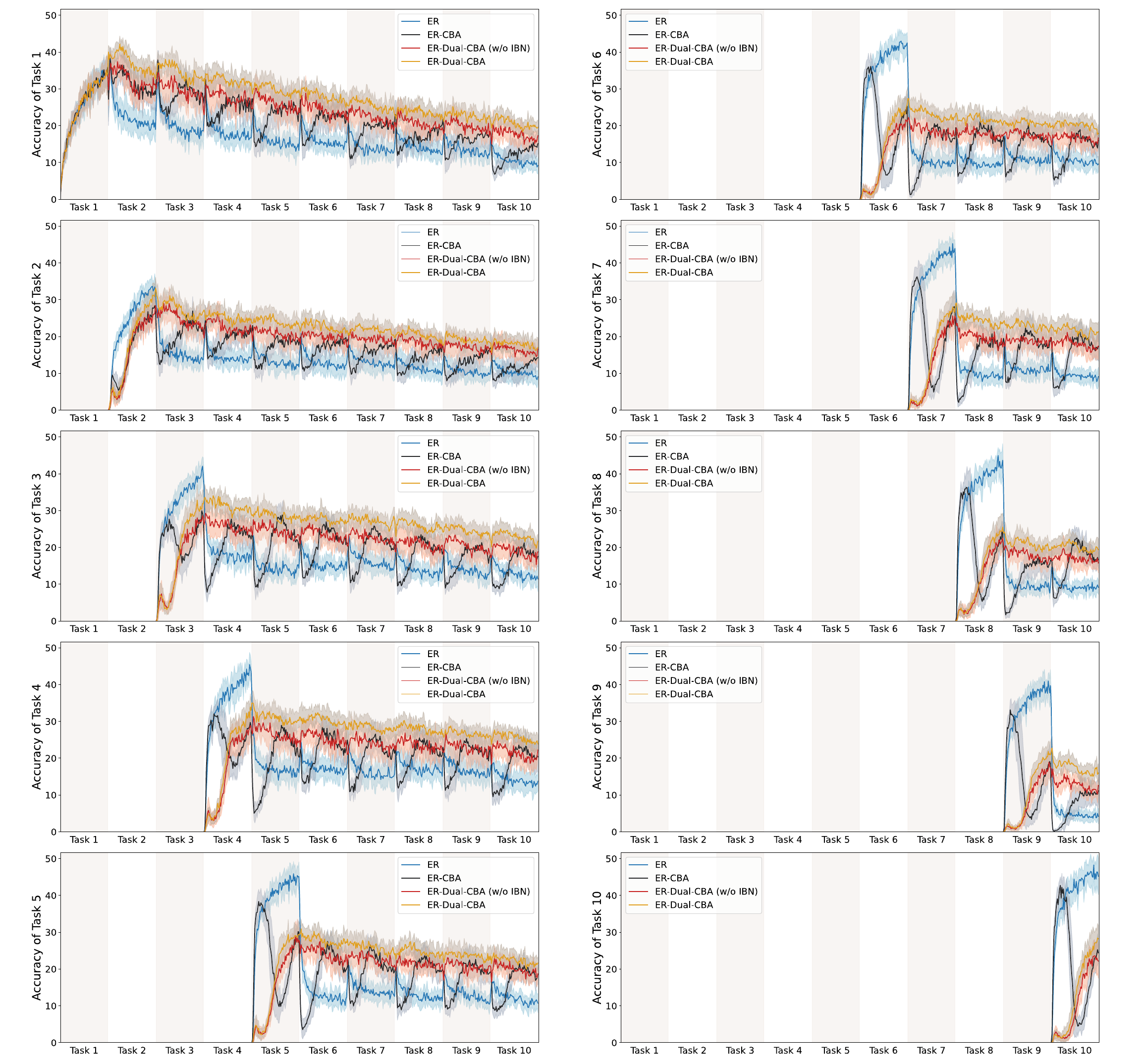}
		\end{center}
		\vspace{-5mm}
		\caption{Illustration of the stability gap problem in continual learning by tracking the accuracy of each task during the whole continual learning on Split Tiny-ImageNet with M=2k.}
		\label{fig:stability_gap}
	\end{figure*}

	\medskip
	\noindent
	\textbf{Details of experiment settings.} Following~\myref{[7]}, as we aforementioned in \myref{Sec.~4.1}, we use ResNet18 as our backbone and optimized by SGD optimizer with a learning rate of 0.03. And the Dual-CBA module is updated by Adam optimizer where the learning rate is 0.001 for Split CIFAR-10 and 0.01 for Split CIFAR-100 and Split Tiny-ImageNet. The batch size is set as 32 in all experiments. Besides, the weight of the logit distillation loss and the rehearsal loss of the baseline DER++~\myref{[7]} are fixed as 0.2 and 0.5, respectively.
	
	\subsection{New and Old Task Posterior Probability Distribution}
	\label{appendix:new_old_logits}
	To further demonstrate the effectiveness of our proposed Dual-CBA module, we illustrate the old-task posterior probabilities $\hat{y}_{old}$ and the new-task posterior probability $\hat{y}_{new}$ predicted by the classification network throughout the entire continual learning process in \myref{Fig.~\ref{fig:distrib_change_tsk_dcba}}, where the classification network is trained by ER-Dual-CBA. As shown in this figure, at all task transition timestamps, the classification network can fairly produce correct probabilities for both old and new tasks and shows a more stable tendency on the relationship between old and new task posterior probabilities as the training progresses.
	These results indicate that our Dual-CBA can significantly facilitate the classification network to learn a stable posterior distribution across all the encountered tasks.

	\subsection{Stability Gap Problem}
	\label{appendix:stability_gap}
	The stability gap indicates the performance of previously learned tasks decreases dramatically and then recovers gradually when new tasks come in. To illustrate the stability gap problem in baseline and the class-specific CBA proposed in our conference version~\myref{[16]}, we track the accuracy of each task throughout the continual learning process in \myref{Fig.~\ref{fig:stability_gap}}. \wqza{For example, when the 5th task comes in, the accuracy of each method is shown in the last figure in the left column of \myref{Fig.~\ref{fig:stability_gap}}. It can be observed that:}
	\begin{itemize}
		\item The baseline ER learns new knowledge of the 5th task quickly and achieves high performance at the end of the 5th task. However, its performance drops dramatically when the subsequent future tasks arrive, demonstrating the stability gap problem in CL.
		\item The class-specific CBA~\myref{[16]} exhibits an `increase-decrease-increase' performance pattern during the training of the 5th task. Initially, the class-specific CBA fails to make correct adjustments for the new task, leading to the classification paying much attention to the 5th task and being interrupted by the stability gap problem, which is the first `increase' stage. This is followed by a decline and subsequent recovery in the performance of the 5th task, which can be attributed to the class-specific CBA gradually focusing on the training bias from the posterior distribution perspective. However, the performance of the 5th task drops significantly and then recovers at each incoming future task in CL, indicating the class-specific CBA exacerbates the stability gap problem.
		\item In contrast, the proposed Dual-CBA (w/o IBN) learns the new knowledge of the 5th task more stably. The performance progressively increases until convergence and keeps maintaining throughout the whole CL process with only a slight decrease.
		\item With the help of IBN, Dual-CBA achieves higher performance and shows a similar trend to Dual-CBA (w/o IBN), indicating that the proposed IBN does not affect the stability gap problem but enhances the overall performance of the classification model.
	\end{itemize}
	In summary, while the CBA proposed in our conference version~\myref{[16]} may seriously aggravate the stability gap problem, our proposed Dual-CBA in this paper significantly assimilates it in the CL process. By capturing the stable relationship between the posterior probability of new and old tasks, Dual-CBA enables stable learning of the knowledge of new tasks with only slight forgetting of previously learned tasks.

	\begin{table*}[]
		\centering
		\caption{Ablation analysis on small buffer sizes.}
		\begin{tabular}{cccccccc}
			\hline
			Dataset                     & Method                              & ACC $\uparrow$                    & FM $\downarrow$                   & ACC $\uparrow$                    & FM $\downarrow$                   & ACC $\uparrow$                    & FM $\downarrow$                   \\ \hline
			&                                     & \multicolumn{2}{c}{M=10}                                              & \multicolumn{2}{c}{M=50}                                              & \multicolumn{2}{c}{M=100}                                             \\ \cline{3-8} 
			& ER                                  & 18.33                             & 77.51                             & 22.44                             & 74.47                             & 28.08                             & 61.30                             \\
			& ER-CBA                              & 18.45                             & 77.23                             & 23.13                             & 71.44                             & 30.91                             & 56.55                             \\
			\multirow{-3}{*}{CIFAR-10}  & \cellcolor{gray!40}ER-Dual-CBA (ours) & \cellcolor{gray!40}\textbf{19.75} & \cellcolor{gray!40}\textbf{56.67} & \cellcolor{gray!40}\textbf{29.70} & \cellcolor{gray!40}\textbf{47.45} & \cellcolor{gray!40}\textbf{36.41} & \cellcolor{gray!40}\textbf{36.94} \\ \hline
			&                                     & \multicolumn{2}{c}{M=100}                                             & \multicolumn{2}{c}{M=200}                                             & \multicolumn{2}{c}{M=500}                                             \\ \cline{3-8} 
			& ER                                  & 8.55                              & 51.18                             & 10.43                             & 48.70                             & 14.43                             & 41.03                             \\
			& ER-CBA                              & 10.16                             & 40.70                             & 13.94                             & 33.95                             & 19.29                             & \textbf{21.83}                    \\
			\multirow{-3}{*}{CIFAR-100} & \cellcolor{gray!40}ER-Dual-CBA (ours) & \cellcolor{gray!40}\textbf{10.79} & \cellcolor{gray!40}\textbf{36.21} & \cellcolor{gray!40}\textbf{14.71} & \cellcolor{gray!40}\textbf{32.12} & \cellcolor{gray!40}\textbf{21.32} & \cellcolor{gray!40}21.90        \\ \hline
			&                                     & \multicolumn{2}{c}{M=200}                                             & \multicolumn{2}{c}{M=500}                                             & \multicolumn{2}{c}{M=1000}                                            \\ \cline{3-8} 
			& ER                                  & 6.37                              & 41.00                             & 8.18                              & 38.19                             & 10.66                             & 33.64                             \\
			& ER-CBA                              & 8.12                              & \textbf{33.84}                    & 10.55                             & 27.24                             & 13.68                             & 17.22                             \\
			\multirow{-3}{*}{TinyImg}   & \cellcolor{gray!40}ER-Dual-CBA (ours) & \cellcolor{gray!40}\textbf{8.89} & \cellcolor{gray!40}34.13         & \cellcolor{gray!40}\textbf{12.90} & \cellcolor{gray!40}\textbf{27.19} & \cellcolor{gray!40}\textbf{17.08} & \cellcolor{gray!40}\textbf{17.13} \\ \hline
		\end{tabular}
		\label{tab:small_buffer}
	\end{table*}

	\subsection{Results of Small Buffer Sizes}
	\label{appendix:small_buffer}
	In \myref{Table I} of the main text, we have reported the results of our method under the three datasets with various buffer sizes and demonstrated the effectiveness of the proposed Dual-CBA module. To further explore the performance of our method under some extreme scenarios, that is, the storage spaces are limited with extremely small buffer sizes in the continual learning process. The comparison results based on ER under small buffer sizes on each dataset are summarized in \myref{Table~\ref{tab:small_buffer}}. Specifically, we gradually reduce the buffer sizes until only one sample is stored per class on average, \textit{i.e.}, for Split CIFAR-10, at least M=10, then for Split CIFAR-100 and Tiny-ImageNet, at least M=100 and 200, respectively. Note that even under these extreme limitations, our proposed Dual-CBA can consistently outperform the baseline ER and ER-CBA for both ACC and FM, demonstrating that the proposed bi-level optimization framework remains effective in alleviating task-recency bias even with small buffer sizes.

	\begin{table}[t]
		\centering
		\caption{Computation and memory on Split CIFAR-10 with M=0.2k. $\dagger$: the average training time for each task.}
		\resizebox{\columnwidth}{!}{
			\begin{tabular}{cccc}
				\hline
				Method       & Params ($\times 10^6$) & GPU memory (Mb) & Training time$^\dagger$ (s) \\ \hline
				ER           & 11.174                 & 1072            & 124.57                      \\
				ER-CBA       & 11.179                 & 1204            & 130.87                      \\
				ER-Dual-CBA  & 11.179                 & 1204            & 134.89                      \\ \hline
		\end{tabular}}
		\label{tab:computation}
		\vspace{-3mm}
	\end{table}

	\subsection{Computation and GPU Memory}
	\label{appendix:gpu_memory}
	\myref{Table~\ref{tab:computation}} lists the training cost on NVIDIA GeForce RTX 4090, and it shows that the proposed Dual-CBA only increases approximately 130~Mb GPU memory and takes no more than 15 seconds extra training time compared with the baselines since it only needs to unroll the gradient of the linear classification layer to update a few hyper-parameters in the outer-loop update. This additional calculation is only a small burden for the CL algorithm and completely meets the online requirement.

}



\vfill

\end{document}